\pdfoutput=1

\documentclass{article}

\PassOptionsToPackage{numbers, compress}{natbib}
\usepackage[final]{neurips_2023}

\usepackage[utf8]{inputenc} % allow utf-8 input
\usepackage[T1]{fontenc}    % use 8-bit T1 fonts
\usepackage{hyperref}       % hyperlinks
\usepackage{url}            % simple URL typesetting
\usepackage{booktabs}       % professional-quality tables
\usepackage{amsfonts}       % blackboard math symbols
\usepackage{nicefrac}       % compact symbols for 1/2, etc.
\usepackage{microtype}      % microtypography
\usepackage[dvipsnames]{xcolor}

\usepackage{times}
\usepackage{graphicx}
\usepackage{tablefootnote}
\usepackage{amsmath}
\usepackage{amssymb}

 \usepackage{tabularx}
\usepackage{longtable}
\usepackage{arydshln}

\def\D{{\mathcal D}}

\def\ty{{\tilde{y}}}

\usepackage{enumitem}
\setlist[itemize]{leftmargin=*}

\usepackage{multirow} % for borders and merged ranges
\usepackage{soul}% for underlines
\usepackage{changepage,threeparttable} % for wide tables
\usepackage{adjustbox}
\usepackage{appendix}
\usepackage{multirow}
\usepackage{subcaption}
\usepackage{comment}
\usepackage{color}

\usepackage{array}

\usepackage{algorithm}
\usepackage{algpseudocode}
\usepackage{xspace}
\usepackage{wrapfig}

\usepackage[symbol]{footmisc}

\newcounter{oldtocdepth}

\newcommand{\hidefromtoc}{%
  \setcounter{oldtocdepth}{\value{tocdepth}}%
  \addtocontents{toc}{\protect\setcounter{tocdepth}{-10}}%
}

\newcommand{\unhidefromtoc}{%
  \addtocontents{toc}{\protect\setcounter{tocdepth}{\value{oldtocdepth}}}%
}

\newcolumntype{C}[1]{>{\centering\let\newline\\\arraybackslash\hspace{0pt}}m{#1}}
\makeatletter
\DeclareRobustCommand\onedot{\futurelet\@let@token\@onedot}
\def\@onedot{\ifx\@let@token.\else.\null\fi\xspace}

\def\etal{\emph{et al}\onedot}
\makeatother

\title{Label-Only Model Inversion Attacks via \\ Knowledge Transfer}

\author{%
  Ngoc-Bao Nguyen$^{*1}$\qquad
  Keshigeyan Chandrasegaran$^{*2 \ddag}$\qquad \\
  \textbf{Milad Abdollahzadeh}$^{1}$\qquad
  \textbf{Ngai-Man Cheung}$^{1 \dag}$\\
  $^1$Singapore University of Technology and Design (SUTD)\qquad
  $^2$Stanford University \and
  \texttt{thibaongoc\_nguyen@mymail.sutd.edu.sg}\quad
  \texttt{ngaiman\_cheung@sutd.edu.sg}
}

\begin{document}

\maketitle
% \footnotetext{$^*$ These authors contributed equally. \quad $^\ddag$ Work done while at SUTD. \quad $^\dag$ Corresponding author.}

\footnotetext{$^*$ These authors contributed equally. \quad $^\ddag$ Work done while at SUTD.}

\footnotetext{$^\dag$ Corresponding author.}

\hidefromtoc

%%%%%%%%% ABSTRACT
\begin{abstract}

In a model inversion (MI) attack, an adversary abuses access to a machine learning (ML) model to infer and reconstruct private training data. Remarkable progress has been made in the white-box and black-box setups, where the adversary has access to the complete model or the model's soft output respectively. However, there is very limited study in the most challenging but practically important setup: Label-only MI attacks, where the adversary only has access to the model's predicted  label (hard label) without confidence scores nor any other model information.  

In this work, we propose LOKT, a novel approach for label-only MI attacks. Our idea is based on transfer of knowledge from the opaque target model to  surrogate models. 
Subsequently, using these surrogate models, our approach can harness advanced white-box attacks. 
We propose knowledge transfer based on generative modelling, and introduce a new model, Target model-assisted ACGAN (T-ACGAN), for effective knowledge transfer. 
Our method casts the challenging label-only MI into the more tractable white-box setup. 
We provide analysis to support that surrogate models based on our approach serve as effective proxies for the target model for MI. Our experiments show that our method significantly \textbf{outperforms existing SOTA Label-only MI attack by more than 15\% across all MI benchmarks.} Furthermore, our method compares favorably in terms of query budget. 
Our study highlights rising privacy threats for  ML models even when minimal information (i.e.,  hard labels) is exposed. Our code, demo, models and reconstructed data are available at our project page:
\textcolor{magenta} {\url{https://ngoc-nguyen-0.github.io/lokt/}}
% \textcolor{magenta} {\href{https://ngoc-nguyen-0.github.io/label_only_tacgan/}{project website}}.

\end{abstract}

\section{Introduction}
\begin{figure*}[!th]
\begin{adjustbox}{width=1.0\textwidth,center}
\begin{tabular}{c}
    \includegraphics[width=0.99\textwidth]{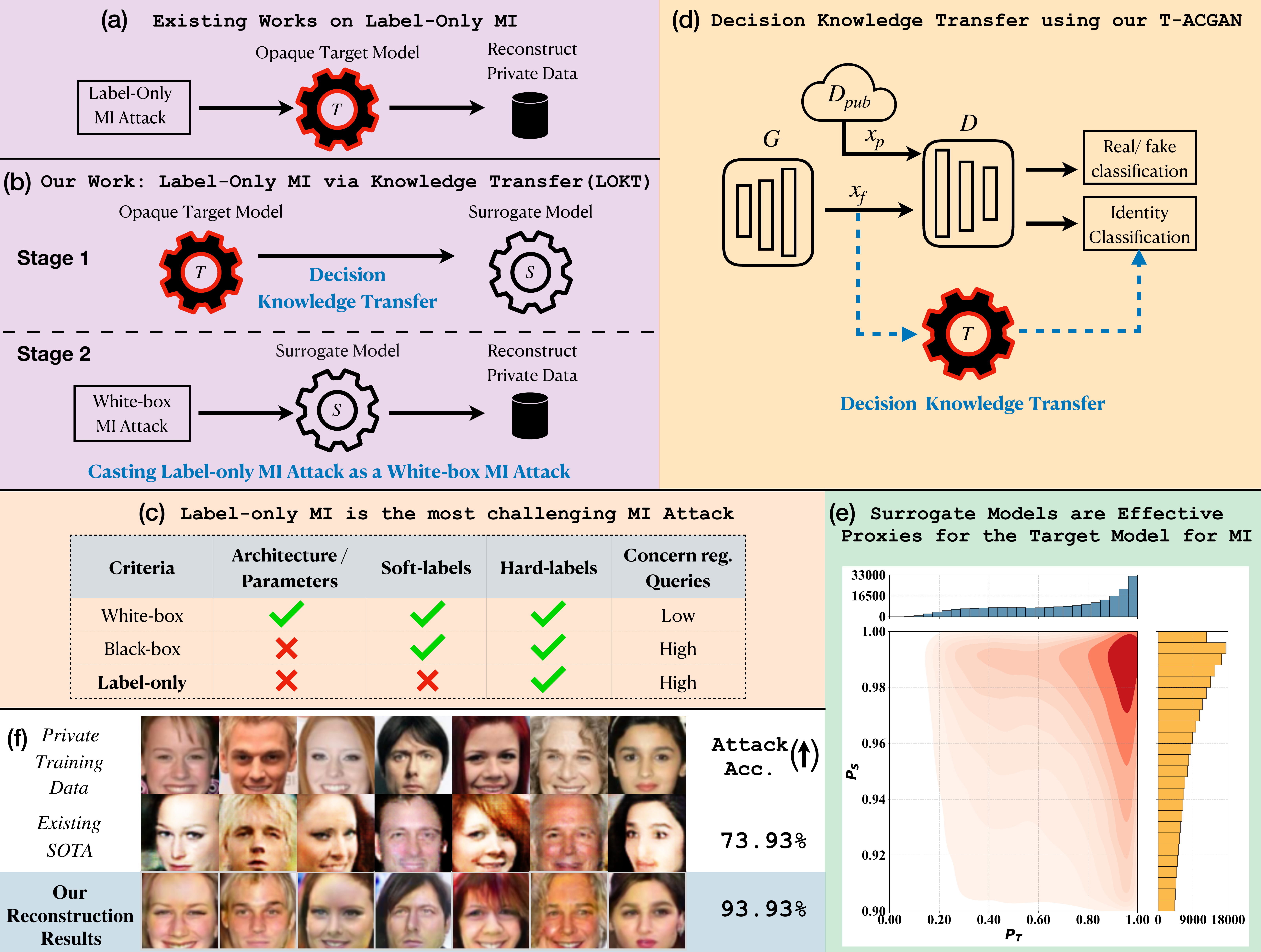}
\end{tabular}
\end{adjustbox}
\vspace{-0.35cm}
\caption{
{\em Overview and our contributions}.
{\bf (a)} Under Label-only model inversion (MI) attack, the Target model $T$ is opaque.
% {\bf (b)} As our first contribution, we propose to cast the Label-only MI attack to a white-box MI attack by training the surrogate model $S$.
{\bf (b) Stage 1:} As our first contribution, we propose a knowledge transfer scheme to render surrogate model(s). 
{\bf (b) Stage 2:} Then, we cast the Label-only MI attack as a white-box MI attack on surrogate model(s) $S$.
{\bf (c)} This casting can ease the challenging problem setup of label-only MI attack into a white-box MI attack. To our knowledge, our proposed approach is the first to address label-only MI via white-box MI attacks.
{\bf (d)} We propose T-ACGAN to leverage generative modeling and the target model for effective knowledge transfer to render surrogate model(s). Knowledge transfer renders $D$ (Discriminator) as a surrogate model, and further generated samples of T-ACGAN can be used to train 
% other
additional 
surrogate variant $S$ (Sec.~\ref{sec:proposed_method}).
{\bf (e)} Our analysis demonstrates that $S$ is an effective proxy for $T$ for MI attack (details in Sec.\ref{sec:analysis}).
In particular, white-box MI attack on $S$ mimics the white-box attack on opaque $T$. 
{\bf (f)} Our proposed approach significantly improves the Label-only MI attack (e.g. $\approx$ 20\% improvement in standard CelebA benchmark compared to existing SOTA \cite{kahla2022label}) resulting in significant improvement in private data reconstruction.
Best viewed in color.
}
\label{fig:framework}
\vspace{-0.5cm}
\end{figure*}

Model inversion (MI) attacks 
aim to infer and reconstruct sensitive private samples used in the training of  models. 
MI and their privacy implications have attracted considerable attention recently
\cite{zhang2020secret,chen2021knowledge,wang2021variational,yang2019neural,pmlr-v162-struppek22a,kahla2022label,yuan2023pseudo,khosravy2022model,wang2021improving, an2022mirror,nguyen_2023_CVPR}.
The model subject to MI is referred to as {\em target model}.
There are three categories of MI attacks: (1) White-box MI, where  complete target model information is accessible by the adversary \cite{zhang2020secret,chen2021knowledge, wang2021variational,pmlr-v162-struppek22a,yuan2023pseudo,an2022mirror}; 
(2) Black-box MI, where target model's soft labels are accessible \cite{yang2019neural,gamin2020blackbox,an2022mirror,Han_2023_CVPR}; (3) Label-only MI, where only target model's hard labels are accessible \cite{kahla2022label}.
This paper focuses on label-only MI, which is the most challenging setup as only limited information (hard labels) 
is available (Fig.~\ref{fig:framework}).

% In most existing work, MI are formulated as optimization problems to seek reconstructions that achieve the highest likelihood under the target model \cite{zhang2020secret,chen2021knowledge,wang2021variational,kahla2022label}.
In most existing work, MI attack is formulated as an optimization problem to seek reconstructions that maximize the likelihood under the target model \cite{zhang2020secret,chen2021knowledge,wang2021variational,kahla2022label}.
For DNNs, the optimization problems are highly non-linear. When the sensitive private samples are high-dimensional samples (e.g. face images), the optimizations are ill-posed, even in white-box setups. 
To overcome such issues, recent MI
\cite{zhang2020secret,chen2021knowledge,wang2021variational,pmlr-v162-struppek22a,an2022mirror,yuan2023pseudo,kahla2022label,nguyen_2023_CVPR}
learn distributional priors from
public data via GANs \cite{goodfellow2014GAN,Karras_2019_CVPR,Karras_2020_CVPR,chandrasegaran2021closer}, and solve the optimization problems
over GAN latent space rather than the unconstrained image
space. 
For example, MI attacks on face recognition systems could leverage GANs to learn face manifolds from public face images which have no identity intersection with private training images.
White-box attacks based on public data and GANs have achieved remarkable success
\cite{zhang2020secret,chen2021knowledge,wang2021variational,yuan2023pseudo,nguyen_2023_CVPR}.
We follow existing work and recent label-only MI \cite{kahla2022label} and leverage public data in our method. Furthermore, similar to existing work, we use face recognition 
% classifiers
models
as examples of target models.

{\bf Research gap.}
Different from white-box attack, study on label-only attack is  limited despite its  practical importance, e.g., many 
practical ML models only expose predicted labels.
Focusing on label-only attack and 
% In 
% \cite{kahla2022label}, the first approach for label-only attack was proposed.
% Specifically, 
with no knowledge of internal workings of target model nor its confidence score,  
BREPMI
\cite{kahla2022label} takes a {\em black-box search} approach  
to explore the search space iteratively
(Fig.~\ref{fig:framework}(a)).
To seek reconstructions with high likelihood under  target model,
\cite{kahla2022label} proposes to query  target model
 and observe the model's hard label predictions, 
 and update 
 search directions 
 using {\em Boundary Repelling} 
 in order to move towards centers of decision regions, where high likelihood reconstructions could be found.
However, 
%Despite their encouraging results, 
black-box search in the high-dimensional latent space is 
% very 
extremely
challenging.

{\bf In this paper}, we propose a new approach for Label-Only MI attack using Knowledge Transfer (LOKT). 
Instead of 
% taking 
performing
a black-box search approach
as demonstrated in \cite{kahla2022label}
and directly searching high-likelihood reconstruction from the opaque target model (Fig.~\ref{fig:framework}(a)), which could be particularly challenging for high-dimensional search space, 
% we propose an approach to transfer the decision knowledge of the target model to surrogate models, for which 
% complete model information is accessible.
we propose a different approach.
Our approach aims to transfer the decision knowledge of the target model to surrogate models, for which 
complete model information is accessible.
% architectures,  parameters and soft outputs are fully accessible. 
Subsequently, with these surrogate models, we could 
harness
% leverage
SOTA white box attacks
to seek high-likelihood reconstructions
(Fig.~\ref{fig:framework}(b)).
To obtain the surrogate models, we explore generative modeling \cite{abdollahzadeh2023survey, zhao2023exploring, teo2023fair, zhao2022few, teo2022cleam}.
% In particular, we propose a new 
% Target model-assisted ACGAN, T-ACGAN, extending ACGAN \cite{odena2017conditional}
% and taking advantage
% of our unique problem setup, i.e., 
% availability of the target model  (even though only predicted labels are accessible in target model)
% (Fig.~\ref{fig:framework}(d)).
In particular, we propose a new 
Target model-assisted ACGAN, T-ACGAN, which extends ACGAN \cite{odena2017conditional}
and leverages our unique problem setup where we have access 
% to the target model 
to the predicted labels of the target model
% (even though only predicted labels are accessible in the target model)
as shown in Fig.~\ref{fig:framework}(d).
In particular, by 
% properly 
effectively
leveraging the target model in discriminator/classifier training, 
we 
% could 
can
explore {\em synthetic data} for decision knowledge transfer from the target model to the surrogate model.
With T-ACGAN
capturing the data manifold
of  public samples, 
synthetic data is diverse and abundant.
We hypothesize that such rich synthetic data could lead to improved decision knowledge transfer. 
% and we hypothesize that we could achieve better decision knowledge transfer. 
% Also, 
Moreover, as training
progresses, T-ACGAN generator learns to improve its conditional generative capabilities,
% to generate more balanced synthetic data,
enabling it to produce more balanced synthetic data
for surrogate model learning.
% We further explore several designs of the surrogate models. 
We explore several surrogate model designs. 
% In one design, we directly take   discriminator and classifier of T-ACGAN as the surrogate model. In another design, we leverage generator of T-ACGAN to train other surrogate model variants.  
In one configuration, we employ the discriminator/ classifier of T-ACGAN as the surrogate model. 
In an alternative design, we utilize the generator of T-ACGAN to train different surrogate model variants.
% Note that generator of T-ACGAN 
% can be readily leveraged for white-box attack, and its conditional generation capability can help reduce the search space during inversion.
It's noteworthy that the generator of T-ACGAN can be readily employed for white-box attacks, and its conditional generation capabilities can effectively reduce the search space during inversion.
% {\bf In addition, we perform analysis to support that 
% our surrogate models are good proxy for the opaque target model for MI.}
{\bf In addition, we perform analysis to support that 
our surrogate models are effective proxies for the opaque target model for MI.}
(Fig.~\ref{fig:framework}(e)).
% Overall, our T-ACGAN can lead to improved surrogate models, which is validated in our ablation experiments, and improved MI attack accuracy 
% (Fig.~\ref{fig:framework}(f))
% and reduced number of query compared to existing SOTA.
Overall, our T-ACGAN renders improved surrogate models, resulting in a significant boost in MI attack accuracy (Fig.~\ref{fig:framework}(f)) and reduced number of queries compared to previous SOTA approach.
{\bf Our contributions are:} 
\begin{itemize}
  \item We propose LOKT, a new label-only MI by  transferring decision knowledge from the target model to surrogate models and performing white-box attacks on the surrogate models (Sec. \ref{sec:approach}).
  Our proposed approach is the first  to address label-only MI via  white-box attacks.
\item
We propose a new T-ACGAN to leverage generative modeling and the target model for effective knowledge transfer (Sec. \ref{sec:approach}).
\item
We perform analysis to support that our surrogate models are 
% good proxy 
effective proxies for the target model for MI (Sec. \ref{sec:analysis}).
\item
We conduct extensive experiments and ablation to support our claims. Experimental results show that our approach can achieve significant improvement compared to existing SOTA MI attacks
(Sec. \ref{sec:experiments}).
Additional experiments/ analysis are in Supplementary.

\end{itemize}

%and we are mindful the number of target model requests in our design

% =================

% Our contribution are:
% \begin{itemize}
%   \item We propose a new method for training shadow models to mimic the decision of the target model based on public data - private label. 
%   \item We propose to cast the challenging label-only MI attack into white-box setup and harnesses SOTA white-box MI attacks to tackle label-only MI. To the best of our knowledge, our proposed method is the first work to address label-only MI via  white-box approaches. 
%   \item Our method significantly outperforms existing SOTA label-only MI attack by more than 10\% across all MI benchmarks.
% \end{itemize}

%%% ==================Main Figure of Penultimate layer visualization =============
\section{Related work}
% In recent times, there has been significant attention directed towards the privacy of deep neural networks (DNNs) \cite{chabanne2017privacy,beaulieu2019privacy,sirichotedumrong2021gan,lee2022privacy,subbanna2021analysis}. 
% represents a crucial class of privacy attacks, which can 
Model Inversion (MI) 
has particularly alarming consequences in security-sensitive domains, such as face recognition \cite{schroff2015facenet, taigman2014deepface, meng2021magface,  huang2020curricularface}, medical diagnosis \cite{dufumier2021contrastive, yang2022towards, dippel2021towards}.
Fredrikson et al.
\cite{fredrikson2014privacy}
introduces the first MI attack
for simple linear regression models.
Recently, several  works extend MI for complex DNNs under different setups.
For white-box setup, 
\cite{zhang2020secret} proposes Generative
Model Inversion (GMI) to leverage public data and GAN\cite{goodfellow2020generative,arjovsky2017wasserstein}
to constrain the search space.
%of high likelihood reconstructions.
\cite{chen2021knowledge} proposes 
Knowledge-Enriched Distributional Model Inversion (KEDMI) to train an  inversion-specific GAN for the attack.
\cite{wang2021variational} proposes 
Variational Model Inversion (VMI) to apply  variational objectives for the attack. 
Very recent work \cite{yuan2023pseudo} proposes  
Pseudo Label-Guided MI (PLG-MI) to apply model's soft output to train a  
 conditional
GAN (cGAN)\cite{miyato2018cgans} for white-box attack. LOMMA\cite{nguyen_2023_CVPR} proposes a better objective function for MI and model augmentation to address MI overfitting.
For black-box attack, where model's soft output is available, 
\cite{yang2019neural} proposes to train an inversion model 
and a decoder to generate target images using predicted scores of the inversion model.
\cite{gamin2020blackbox} proposes an adversarial approach for black-box MI.
For label-only attack, \cite{kahla2022label} proposes BREPMI, the first label-only MI
using a  black-box Boundary Repelling search.
See Supplementary for further discussion of related work.

\section{Problem setup}
\label{sec:problemSetup}

Given a target model $T$, the goal of MI is to infer private training data $\mathcal{D}_{priv}$ by abusing access to model $T$. 
More specifically, given a target class/ identity label $y$, 
the adversary aims to reconstruct an image $x$ 
which is similar to the images of class $y$ in $\mathcal{D}_{priv}$. 
Most MI formulate the inversion as optimization problems to seek the highest likelihood reconstructions 
for identity $y$
under $T$.
As direct  searching for $x$ in the unconstrained image space
is ill-posed, many  MI attacks 
\cite{zhang2020secret,chen2021knowledge,wang2021variational,yuan2023pseudo,kahla2022label}
leverage 
public dataset 
$\mathcal{D}_{pub}$ that is the same domain as $\mathcal{D}_{priv}$, e.g., $\mathcal{D}_{priv}$ and $\mathcal{D}_{pub}$ are  facial image datasets.
GAN \cite{goodfellow2014GAN} is applied to learn 
distributional prior from $\mathcal{D}_{pub}$, and
 the adversary searches the GAN latent space 
instead of the  unconstrained image space
 for high-likelihood reconstructions under $T$:
\begin{equation}
\label{eq:GMI_general}
 %z^* = \arg 
 \max_z \log P_{T} (y|G(z))       
 %\mathcal{L}_{id}(x;y,T) 
\end{equation}
Here, $G$ is the generator, and $P_{T} (y|.)$ is the likelihood of an input for identity $y$ under target model $T$.
White-box attacks 
apply gradient ascent and some regularization 
\cite{zhang2020secret,chen2021knowledge,wang2021variational,yuan2023pseudo}
to solve Eq. \ref{eq:GMI_general},
whereas label-only attack 
BREPMI \cite{kahla2022label} applies black-box search to tackle  
Eq. \ref{eq:GMI_general}.
In this paper, we  also tackle  
Eq. \ref{eq:GMI_general} under label-only setup, i.e. only the predicted label
is available.

\section{Approach}
\label{sec:approach}

Our proposed label-only MI consists of two stages. In stage 1, we learn surrogate models. In stage 2, we apply SOTA white-box attack on the surrogate models. 
To learn surrogate models, we explore 
an approach based on GAN and propose a new Target model-assisted ACGAN (T-ACGAN) for effective  transfer of decision knowledge.
Our T-ACGAN learns the generator $G$ and the discriminator $D$ with classifier head $C$. 
In one setup, we directly take $C \circ D$  as the surrogate model\footnote{With a slight abuse of notation we use $D$ to represent  the entire discriminator and the discriminator up to and including the penultimate layer in the context of  $C \circ D$.}.
In another setup, we apply $G$ to generate synthetic data to train another surrogate model 
$S$ or an ensemble of $S$.
Then, we apply SOTA white-box attack on
$C \circ D$, 
$S$ or the ensemble of $S$. 
In our experiments, we show that using  
 $C \circ D$ in a white-box attack can already outperform existing SOTA label-only attack. Using $S$ or an ensemble of $S$ can further improve  attack performance.
The $G$ obtained from our T-ACGAN can be readily  leveraged in the attack stage.

% {\bf In Sec. \ref{sec:analysis}, we will justify our surrogate model as a good proxy for the opaque target model for inversion.}

\subsection{Baseline}
\label{ssec:baselineForSurrogate}
Before discussing our proposed approach, 
we first discuss a simple baseline for comparison. 
Given the public data, one could directly use the target model $T$ to label the data and learn the 
surrogate model $S$.
For $x_p \in \mathcal{D}_{pub}$, 
we construct 
$ (x_p, \tilde{y}) $,
%$\{ (x_p, \tilde{y}) \}$, 
where 
$\tilde{y} = T(x_p)$ is pseudo label of {\em private} identity.
We obtain the dataset 
$ \tilde{\D}_{pub}$ with samples  $(x_p, \tilde{y})$, 
% $ \tilde{\D}_{pub} = \{ (x_p, \tilde{y}) \}$, 
i.e. $\tilde{\D}_{pub}$ is the public dataset with pseudo labels.
We apply $\tilde{\D}_{pub}$ to train $S$. However, this algorithm suffers from class imbalance. In particular,  
some private identities could have less resemblance to 
%$x_p \in \mathcal{D}_{pub} $. 
$x_p \in \D_{pub} $. 
As a result, for some $\tilde{y}$, there is only a small number of $x_p$ classified into it, 
%by $T$, and 
and 
 $\tilde{\D}_{pub}$ is class imbalanced.
When  using 
$\tilde{\D}_{pub}$ to train $S$, minority classes 
% (private identity 
% $\tilde{y}$ with small amount of $x_p$ classified into) 
may not gain adequate decision knowledge under $T$ and 
could perform sub-optimally.
%, impacting the attack performance on those classes. 
%Therefore, as a variation of Algorithm-A, 
In our experiments, we also apply  techniques to mitigate the class imbalance in  $\tilde{\D}_{pub}$. However, the performance of this baseline approach is inadequate as we will show in the experiments.

\subsection{Review of ACGAN}
In standard ACGAN \cite{odena2017conditional}, we are given a real training dataset with label, i.e., 
$\D_{real}$ with samples
$(x_r, y)$.
%$\{(x_r, y)\}$.
The  generator $G$ takes a random noise vector $z$ and a class label $y$ as inputs to generate a fake sample $x_f$. The discriminator $D$ outputs both a probability distribution over sources $P(s|x) = D(x)$, 
where $s \in \{ Real, Fake \}$, 
and a probability distribution over the class labels, i.e.,  $P(c|x)   =  C \circ D(x)$,  and $c$ is one of the classes.
For real training sample $x_r$ of label $y$ and fake sample $x_f = G(z,y)$ with conditional information $y$, the loss functions for  $D$, $C$ and $G$
are: 
\begin{equation}
 \begin{aligned}
 \mathcal{L}_{D,C} & =  - E[\log P(s = Fake| x_{f})] - E[\log P(s = Real| x_r)] \\&   ~~~~      - E[\log P(c = y | x_f)] - E[\log P(c = y | x_r) ]
\end{aligned}
\label{eq:acgan_D}
\end{equation}
\begin{equation}
 \mathcal{L}_G = E[\log P(s = Fake| x_{f})] -  E[\log P(c = {y} | x_f)]
 \label{eq:acgan_G}
\end{equation}

% \textcolor{red}{\bf [Milad: as $D(x)$ is used for real/fake classification, we may denote the penultimate layer of $D$ as $D^{'}$, and define the output over classes as $C \circ D^{'}(x)$.]}

\subsection{Our Proposed  T-ACGAN and Learning of Surrogate Model}
\label{sec:proposed_method}
Unlike  standard ACGAN setup where
we have access to labelled data  
$\D_{real}$ with samples
$(x_r, y)$,
in our setup, we have access to real 
public data without label: 
%$\mathcal{D}_{pub} = \{ x_p \}$
$\D_{pub}$ with samples $x_p$.
% $x_p \in \mathcal{D}_{pub}$ without label.
{\em Importantly, we can leverage the target model $T$ to provide pseudo labels for  
generated samples 
$x_f = G(z,y)$, which are diverse and abundant.}
Our proposed T-ACGAN aims to take advantage of $T$ to provide more  diverse and accurate pseudo labelled 
samples during the training.

%to learn $D$, $C$ and $G$ better.

% After the training, we take $C \circ D$ as the surrogate model for white-box attack.

{\bf {\boldmath $D$} and {\boldmath $C$} Learning.} 
% For $x_p \in \mathcal{D}_{pub}$, we assign a pseudo label of {\em private} identity:
% $\tilde{y} = T(x_p)$.
Our T-ACGAN  leverages  $T$ to assign pseudo labels to the diverse generated samples $x_f = G(z,y)$, i.e.,   $\tilde{y} = T(x_f)$.
We apply samples 
$x_p$
%$\{x_p\}$
% and 
% $\{(x_p, \ty)\}$ and 
and 
$(x_f, \ty)$ 
%$\{(x_f, \ty)\}$ 
to learn 
$D$ and $C$:
\begin{equation}
 \begin{aligned}
 \mathcal{L}_{D,C} & =  - E[\log P(s = Fake| x_{f})] - E[\log P(s = Real| x_p)] \\&   ~~~~      - E[\log P(c = \ty | x_f)] 
% - E[\log P(c = \ty | x_p) ]
\end{aligned}
\label{eq:tacgan_D}
\end{equation}

% \begin{equation}
%  \begin{aligned}
%  \mathcal{L}_D & =  - E[\log P(s = Fake| x_{f})] - E[\log P(s = Real| x_p)] \\&   ~~~~      - E[\log P(c = \ty | x_f)] - E[\log P(c = \ty | x_p) ]
% \end{aligned}
% \label{eq:tacgan_D}
% \end{equation}
% where $x_f = G(z,y)$, $\tilde{y} = T(x_f)$, and 
% $x_p\in \mathcal{D}_{pub}$.

In Eq.~\ref{eq:tacgan_D}, the  term $E[\log P(c = \ty | x_f)] =
E[\log P(c = \ty | G(z,y))]$ 
is different from ACGAN and may look intriguing. Instead of using $y$ as class supervision to train $D$ and $C$ as in ACGAN 
(Eq.~\ref{eq:acgan_D}), our T-ACGAN takes advantage of $T$ to apply
$\ty = T(G(z,y))$
to train $D$ and $C$, as $\ty$ is more accurate conditional information compared with $y$ especially during the initial epochs. 
{\em 
With Eq.~\ref{eq:tacgan_D}, our method transfers the decision knowledge of $T$ into  
 $D$ and $C$ via diverse generated samples.} 
% We remark that in standard ACGAN, conditional signal $y$ of $x_f=G(z,y)$ is used as class supervision to train $D$ and $C$. 
% However, in our setup, since we have access to $T$, we could use more accurate $\ty = T(G(z,y))$ as class supervision to train $D$ and $C$ (compare Eq. \ref{eq:acgan_D} and Eq. \ref{eq:tacgan_D}).
% This is important as $G(z,y)$ may not  generate accurate sample of class $y$ especially at the beginning of the training. 
Furthermore, as we can generate  diverse 
pseudo labelled samples
$(x_f, \ty)$
%$\{(x_f, \ty)\}$ 
using $T$ and $G$, pseudo labelled data based on  
$x_p$
%$\{x_p\}$ 
can be omitted.
% $\{(x_p, \ty)\}$ is not critical. In our experiments, we will show that we can use $\{(x_f, \ty)\}$ alone for class supervision, and the last  term associated with $\{(x_p, \ty)\}$ in Eq. \ref{eq:tacgan_D} could be omitted. 
In our experiment, we show that we can achieve good performance using  diverse samples  
$(x_f, \ty)$.
%$\{(x_f, \ty)\}$. 
In  T-ACGAN, we utilize
public data $x_p$
%$\{x_p\}$ 
only for real/fake discrimination.

{\bf {\boldmath $G$} Learning.} 
We follow ACGAN training for $G$, i.e. Eq.~\ref{eq:acgan_G}.
With $D$ and $C$ trained with decision knowledge of $T$ in the above step, they provide feedbacks to $G$  to improve its conditional generation
{\em in the private label space of $T$}.
In our experiment, we analyze  $y$ in $x_f = G(z,y)$ and  
$\tilde{y} = T(x_f)$. 
As training progresses, $G$  improves its conditional generation, and $y$ and $\ty$ become more aligned.
Note that, as $T$ outputs only hard labels,  $T$ cannot be readily applied to provide feedback for $G$ learning.

{\bf 
Surrogate Model.}
With alternating $D$ and $C$ learning and $G$ learning, we obtain $D$, $C$ and $G$.
We explore three methods to obtain the surrogate model. 
$\bullet$ (i) We directly take 
$C \circ D$ 
in T-ACGAN
as the surrogate model and apply  a white-box attack on $C \circ D$.
This can be justified as $C \circ D$ is trained based on decision knowledge of $T$ to classify a sample into identities of private training data. $\bullet$ (ii) We apply $G$ of T-ACGAN to generate
dataset 
$\tilde{\D}_{fake}$
with samples 
$(x_f, \ty)$,
%$\{(x_f, \ty)\}$, 
where 
$x_f = G(z,y)$ and $\tilde{y} = T(x_f)$.
We apply
$\tilde{\D}_{fake}$
%$\{(x_f, \ty)\}$ 
to train another surrogate model $S$.
% Note that while $y$ and $\ty$ are well aligned as shown in Figure, $\ty$ remains more accurate as it is the output from $T$.
% Therefore, we apply $\ty$ as supervision to train $S$.
$\bullet$
(iii) We use the same dataset 
$\tilde{\D}_{fake}$
%$\{(x_f, \ty)\}$ 
in (ii) to train an ensemble of $S$ of different architectures.
As pointed out in
\cite{nguyen_2023_CVPR}, 
using an ensemble of $S$ could improve white-box attack performance.
% {\bf In Sec., we justify why our surrogate model is a good proxy for the opaque target model in MI.}

{\bf White-box Attack.}
With surrogate model $C \circ D$, $S$ or an ensemble of $S$, any white-box attack can be applied.
In our experiments, we show that our surrogate models are effective across a range of white-box attacks (See the Supplementary).
Furthermore, 
$G$ in T-ACGAN can be readily leveraged for performing the attack.
Particularly, based on $G(z,y)$ obtained in the above steps,
we could  reduce the search space during inversion to the latent region 
corresponding to the target identity $y$, leading to more efficient search and improved attack accuracy \cite{yuan2023pseudo}.

\section{Analysis for justification of surrogate models}
\label{sec:analysis}

In this section, we provide an analysis to justify why
our surrogate model 
could be 
% a good proxy
an effective proxy
for $T$ under MI, i.e., {\em the results of white-box MI attack on our surrogate model  be good approximation to that of white-box MI attack on $T$}.
Note that results of white-box MI on $T$ cannot be obtained directly as $T$ exposes only hard labels.
To simplify the presentation, we focus our discussion on $S$.
As discussed in Sec. \ref{sec:problemSetup}, 
most MI attacks formulate inversion  as an optimization problem of seeking reconstructions that achieve highest likelihood under  target model.
Therefore, when we carry out MI on $S$ with SOTA white-box approaches, we expect to obtain high-likelihood reconstructions under $S$ (or high-likelihood generated samples of GAN under $S$, see Eq. \ref{eq:GMI_general}).
We use $P_S$ and $P_T$ to denote likelihood of a sample  under $S$ and $T$ respectively.

% ; we remark that every reconstruction obtained via generative MI is a generated sample of some GAN).

In what follows, we provide analysis to support that $S$ based on our approach would possess an important property of good proxy for $T$.
% $\bullet$
    \textcolor{magenta}{\bf Property P1:} 
    {\em 
    For high-likelihood samples under $S$, it is likely that they 
    also have high likelihood under $T$.}
    See Fig. \ref{fig:framework}(e) for distribution of generated samples' $P_T$  conditioning on those with high $P_S$. It can be observed that many  have high $P_T$.
    %many samples in region of high likelihood under $S$ and $T$).
% $\bullet$
    %\textcolor{magenta}{\bf Property P2:} 
    Particularly, it is uncommon for high-likelihood samples under $S$ to have low likelihood under $T$ (see Fig. \ref{fig:framework}(e) only a few samples have low $P_T$).

With
\textcolor{magenta}{\bf Property P1},
the result obtained by white-box on $S$ (which is a high likelihood sample under $S$) is likely to have a high likelihood under $T$ and could be a good approximation to the result of white-box on $T$ (which is a high likelihood sample under $T$). In Fig. \ref{fig:framework}(e), 
%we show the distribution of the likelihood of generated samples under $T$ and $S$. 
\textcolor{magenta}{\bf P1}  can be clearly observed\footnote{Fig. \ref{fig:framework}(e) are $P_T$ and $P_S$ of $x_f= G(z,y)$ from our T-ACGAN. More details in Supp.}.
Therefore, $S$ using our approach would possess \textcolor{magenta}{\bf P1} and would be a good proxy for $T$ for MI.

{\bf Why would {\boldmath $S$} possess property \textcolor{magenta}{P1}?} This could be intriguing. After all, $T$ does not expose any likelihood information.  The labels of samples assigned by $T$ are the only information available to $S$ during training of $S$. It does not appear that $S$ can discern low or high-likelihood samples under $T$.

To discuss why $S$ would possess
\textcolor{magenta}{\bf P1}, we apply findings and analysis framework of Arpit et al. \cite{arpit2017} regarding the general learning dynamics of DNNs. \cite{arpit2017} presents a data-centric study of DNN learning with SGD-variants. In \cite{arpit2017}, ``easy samples'' are ones that fit better some patterns in the data (and correspondingly ``hard samples''). The easy and hard samples exhibit high and low likelihoods in DNNs resp. 
%(and small and high {\em loss-sensitivity} resp.) 
as discussed in 
\cite{arpit2017}.
Furthermore, 
an important finding from \cite{arpit2017} is that, in DNNs learning, the models learn simple and general patterns of the data {\em first} in the training stage to fit the easy samples.

\begin{figure}[t]
% \vspace{-0.5cm}
\begin{adjustbox}{width=1.0\textwidth,center}
\begin{tabular}{c}
    \includegraphics[width=0.99\textwidth]{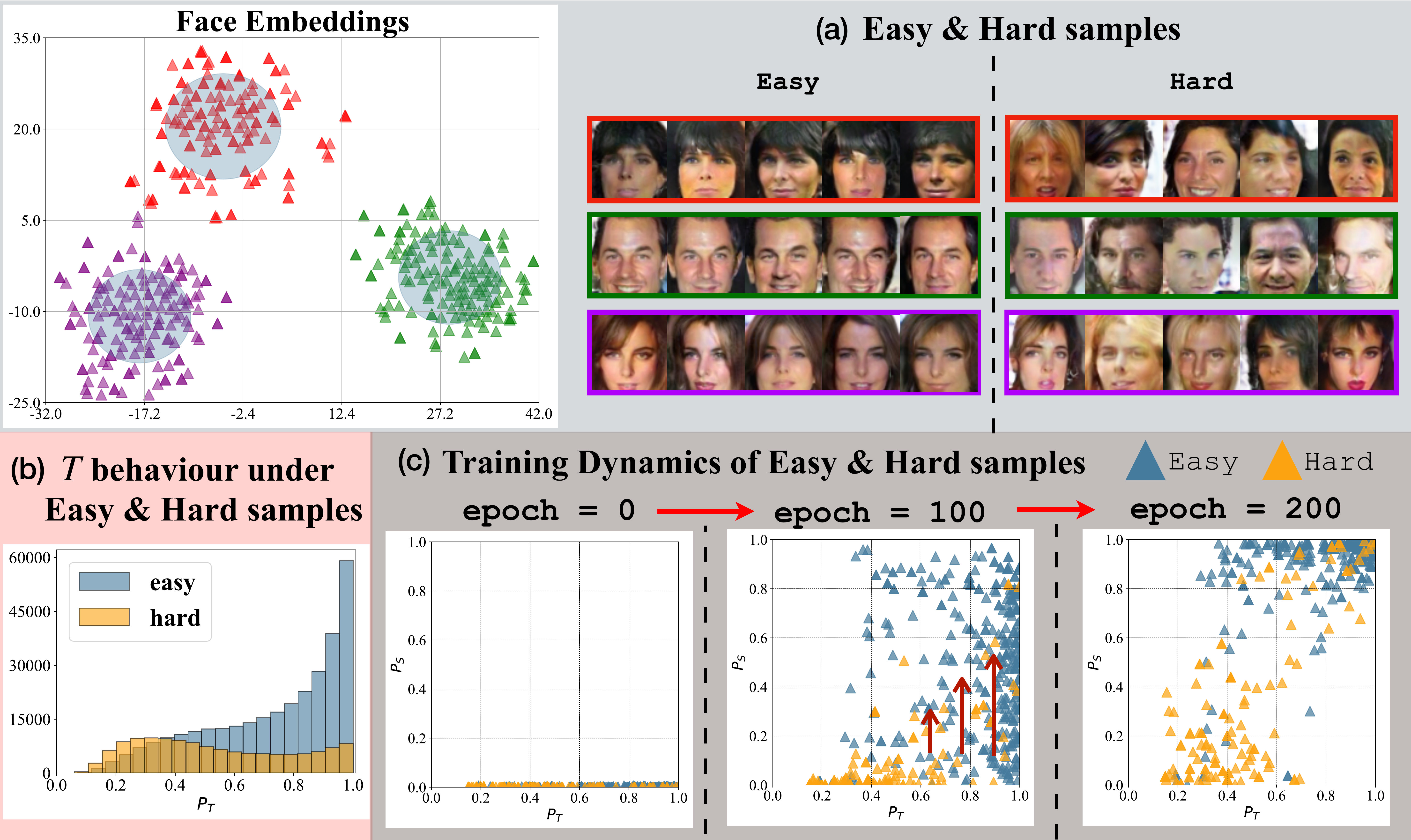}
\end{tabular}
\end{adjustbox}
\vspace{-0.35cm}
\caption{
% {\color{red} Man: editing ===========}
We apply the framework of \cite{arpit2017} to analyze 
learning dynamics of $S$ to reason why $S$ possesses property \textcolor{magenta}{P1}, 
and therefore could be 
% good proxy
an effective proxy for $T$ under MI. 
We analyze generated samples $x_f$ from our T-ACGAN for 3 identities (IDs {\color{red} 20}, {\color{ForestGreen} 16}, {\color{Plum} 36}). 
Note that $x_f$ analysis is relevant as generated samples are used in MI attacks.
{\bf (a)}: 
We analyze face embeddings of $x_f$ extracted from publicly available SOTA face recognition model \href{https://github.com/ageitgey/face_recognition}{here}.
% \footnote{adf}.
% % https://github.com/ageitgey/face_recognition
% {\color{red} color: footnote}
Different clusters and different distances from cluster centroids can be observed, suggesting 
patterns of face identities in some samples (easy samples) while
diverse
appearance in 
other samples (hard samples). 
We use distances from centroids to identify easy samples  
$x_f^e$ and hard samples $x_f^h$ (easy samples are indicated using transparent blue circle for each ID in the visualization). 
Visualization of $x_f^e$ and $x_f^h$ in image space further demonstrates identity patterns in $x_f^e$
and diverse
appearance in $x_f^h$.
{\bf (b)}:
Similar to \cite{arpit2017}, we observe that 
$x_f^e$ and $x_f^h$ tend to have high and low likelihood under $T$ ($P_T$) resp (training data).
{\bf (c)}:  
We track likelihood under $S$ ($P_S$) for 
$x_f^e$ and $x_f^h$ during the training of $S$.
As training progresses, $P_S$ of $x_f^e$ and $x_f^h$ improve, and samples move up vertically (note that $P_T$ of samples do not change).
Consistent with the 
{\em 
``DNNs Learn Patterns First''}
finding
in \cite{arpit2017}, $S$ learns general identity patterns first to fit the easy samples.
Therefore, $P_S$ of $x_f^e$ improve at a faster pace in the training, and many of them achieve high $P_S$
at epoch = 200.
As $x_f^e$ tend to have high $P_T$, we observe property \textcolor{magenta}{P1} in $S$.
% , achieves high likelihood on the easy samples (which also exhibit high likelihood under $T$). Hence we observe property P1 in $S$. 
For $x_f^h$ (many of them tend to have low $P_T$), 
%which exhibit low likelihood under $T$), 
it is uncommon for $S$ to achieve high likelihood on them as they do not fit easily to the pattern learned by $S$. 
{\bf See Supplementary for additional details and  analysis.}
Best viewed in color.
}
\label{fig:analysis}
\vspace{-0.5cm}
\end{figure}

We apply the framework of \cite{arpit2017} to understand our learning of $S$ and the reason why $S$ would possess \textcolor{magenta}{\bf P1}.
Fig. \ref{fig:analysis}(a)
illustrates easy and hard samples in our problem: patterns of face identities can be observed in some samples (easy samples), while other samples (hard samples) exhibit diverse appearance.
Similar to \cite{arpit2017},
Fig. \ref{fig:analysis}(b) shows that these easy and hard face samples tend to have  high and low likelihood under $T$.
Fig. \ref{fig:analysis}(c) shows the learning of $S$ on these easy and hard samples at different epochs. 
Consistent with the 
{\em 
``DNNs Learn Patterns First''}
finding
in \cite{arpit2017}, $S$ learns general identity patterns first to fit the easy samples.
Therefore, $P_S$ of easy samples improve at a faster pace in the training, and many of them achieve high $P_S$. As easy samples tend to have high $P_T$, we observe 
\textcolor{magenta}{\bf P1}
in $S$.
% , achieves high likelihood on the easy samples (which also exhibit high likelihood under $T$). Hence we observe property P1 in $S$. 
For the hard samples (which tend to have low $P_T$), 
%which exhibit low likelihood under $T$), 
it is uncommon for $S$ to achieve high likelihood on them as they do not fit easily to the pattern learned by $S$. 

%Therefore, we observe property P2 in $S$.

% Keshik
%% ==================Main Figure of Penultimate layer visualization =============

%%% ==================End of Penultimate layer visualization =============

\section{Experiments}
\label{sec:experiments}
% In this section, we discuss our extensive experiments and ablation:
% $\bullet$
% (i) We show that our proposed T-ACGAN can lead to better surrogate models compared with alternative approaches (Sec. \ref{sec:surrogate_model}).
% % include model stealing 
% $\bullet$
% (ii) We show that the proposed approach LOKT can outperform the existing SOTA label-only MI attack (Sec. \ref{sec:main_results}).
% $\bullet$
% (iii) We present additional results (Sec. \ref{sec:additional_results}) to demonstrate the efficacy of our surrogate models 
% % with a range of white-box MI attacks, better performance of the proposed method under 
% with the MI defense models, and favorable query budget of the proposed method compared to existing work. 
%Our approach outperforms the existing SOTA label-only MI attack under the MI defense model, and our method compares favorably in terms of query budget.

In this section, we present extensive experiment results and ablation studies:
(i) We show that our proposed T-ACGAN can lead to better surrogate models compared to alternative approaches (Sec. \ref{sec:surrogate_model}).
% include model stealing 
(ii) We show that our proposed approach LOKT can significantly outperform the existing SOTA label-only MI attack (Sec. \ref{sec:main_results}).
(iii) We present additional results (Sec. \ref{sec:additional_results}) to demonstrate the efficacy of LOKT against SOTA MI defense methods. 
We further show that LOKT compares favorably in terms of query budget compared to existing SOTA. 
{\bf Additional experiments/analysis provided in Supplementary.}

\subsection{Experimental Setup}

\begin{wraptable}{r}{5cm}
\vspace{-0.5cm}
\caption{Details of target model $T$. To showcase the effectiveness of our proposed method, we conduct a comprehensive set of 30 experiments, covering 10 different setups.
%Target model $T$ and evaluation model $E$ are provided by \cite{chen2021knowledge,kahla2022label,peng2022bilateral}.
}
\begin{adjustbox}{width=5cm,center}
\begin{tabular}{llc}
\hline
% \boldmath{$\D_{priv}$} & \boldmath{$T$} & \textbf{\# classes} \\ \hline

\multicolumn{1}{c}{\multirow{2}{*}{\boldmath{$\D_{priv}$}}} & \multicolumn{2}{c}{\boldmath{$T$}} \\ \cmidrule(lr){2-3}
  & \textbf{Architecture} & \textbf{\# classes}  \\ \hline

\multirow{5}{*}{CelebA} & FaceNet64 \cite{cheng2017know} & \multirow{5}{*}{1,000}   \\
 & IR152 \cite{he2016deep} &  \\
 & VGG16 \cite{simonyan2014very}&  \\
 & BiDO-HSIC \cite{peng2022bilateral} & \\
  & MID \cite{wang2021improving} & \\ \hline
Facescrub & FaceNet64 \cite{cheng2017know} & 200   \\ \hline
Pubfig83 & FaceNet64 & 50  \\ \hline
\end{tabular}
\end{adjustbox}
\label{tab:target_eval_classifier}
\vspace{-0.8cm}
\end{wraptable}

% To ensure a fair comparison, when evaluating our method, we use the exact same experimental setup as BREPMI \cite{kahla2022label}.
To ensure a fair comparison, we adopt the exact experimental setup used in BREPMI \cite{kahla2022label}.
In what follows, we provide details of the experimental setup.
%Furthermore, we conduct the experiment on the SOTA defense model BiDO-HSIC \cite{peng2022bilateral} which is trained to defend again MI attacks.

\textbf{Dataset.}
We use three datasets, namely CelebA \cite{liu2015deep}, Facescrub \cite{ng2014data}, and Pubfig83 \cite{pinto2011scaling}. 
% We further examine the distribution shift by using FFHQ dataset \cite{karras2019style} which includes images that vary in terms of background, ethnicity, and age. 
We further study Label-Only MI attacks under distribution shift using FFHQ dataset \cite{karras2019style} which contains images that vary in terms of background, ethnicity, and age. 
Following \cite{chen2021knowledge,kahla2022label}, we divide each dataset (CelebA/ Facescrub/ Pubfig83) into two non-overlapping sets: private set $\D_{priv}$ for training the target model $T$, and public set $\D_{pub}$ for training GAN/T-ACGAN.
More details on datasets and attacked identities can be found in Supplementary.

\textbf{Target Models.} Following \cite{chen2021knowledge,kahla2022label}, we use 3 target models $T$ including VGG16 \cite{simonyan2014very}, IR152 \cite{he2016deep}, and FaceNet64 \cite{cheng2017know}. 
All target models are provided in \cite{chen2021knowledge,kahla2022label}. 
%We further compare the attack results under the defense MI attack model. Specially, 
Additionally, we use the following methods/ models for evaluating the attack performance under SOTA MI defense methods:
$\bullet$ BiDO-HSIC \cite{peng2022bilateral}\footnote{
% The pre-trained models are available at 
\url{https://github.com/AlanPeng0897/Defend_MI}}.
$\bullet$ MID \cite{wang2021improving}\footnote{
% We use the official implementation at
\url{https://github.com/Jiachen-T-Wang/mi-defense} 
% to train MID on CelebA setup obtaining natural accuracy of 79.16\%
}.
% pre-trained MI defense model 
% BiDO-HSIC, provided by \cite{peng2022bilateral}. 
% For the defense model MID \cite{wang2021improving}, we use the the official implementation to train MID on CelebA setup (Natural Acc.=79.16\%).
The details are included in Table \ref{tab:target_eval_classifier}.

\textbf{Evaluation Metrics.} 
%To evaluate the efficacy of MI attacks, 
% Following \cite{kahla2022label,nguyen_2023_CVPR,chen2021knowledge}, we employ two metrics to ascertain the extent to which the reconstructed images divulge sensitive information pertaining to the target identity:
Following \cite{kahla2022label,nguyen_2023_CVPR,chen2021knowledge}, we use the following metrics to quantitatively evaluate the performance of MI attacks. Further, we also conduct user studies to assess the quality of reconstructed data (Sec. \ref{main-sub-sec:user-studies}).

% In order to assess the effectiveness of MI attacks, we use two metrics to determine whether the reconstructed images reveal private information related to the target identity:
% it is necessary to determine whether the reconstructed image reveals any confidential information related to a particular label or identity. To accomplish this, our approach follows standard practices in the field by conducting both qualitative evaluations, such as visual inspection, and quantitative evaluations using various metrics, including:
\begin{itemize}[noitemsep,nolistsep]
 \item \textbf{\em Attack Accuracy (Attack acc.)}: 
 % Consistent with 
 Following
 % previous works 
 \cite{zhang2020secret, chen2021knowledge, kahla2022label}, we utilize an evaluation model, 
 % denoted as 
 $E$, which employs a distinct architecture and is trained on  $\D_{priv}$ \footnote{Following previous work, $E$ can also be trained on $\D_{priv}$ and samples from additional identities. This could improve generalization of $E$ for accurate evaluation.}. 
 $E$ serves as a proxy for human inspection \cite{zhang2020secret}. 
 % It is responsible for classifying the reconstructed images. 
 Higher attack accuracy 
 % on the reconstructed images 
 indicates superior performance.
 % Following \cite{zhang2020secret, chen2021knowledge, kahla2022label}, we use an evaluation model $E$ which have different architecture and trained on a superset of $\D_{priv}$ (see Table \ref{tab:target_eval_classifier}). The evaluation model $\mathnormal{E}$ plays a role as a proxy for human inspection \cite{zhang2020secret} to classify the reconstructed images. High accuracy on reconstructed images is better.
 
 \end{itemize}
 
 \begin{itemize}[noitemsep,nolistsep]
 \item \textbf{\em KNN Distance (KNN dt.)}: The KNN distance indicates the shortest distance between the reconstructed image of a specific identity and its private images. Specifically, the shortest distance is computed using the $l_2$ distance in the feature space, using the evaluation model's penultimate layer. A smaller KNN distance signifies that the reconstructed images are more closely aligned with the private images of the target identity.
\end{itemize}

\subsection{Training surrogate model with different algorithms}
\label{sec:surrogate_model}

In this section, we demonstrate that our proposed T-ACGAN can lead to  better surrogate models for MI. 
% For this purpose, 
We describe a set of alternative approaches that can be used to train surrogate models using $\D_{pub}$ and compare the performance of these approaches 
with our proposed method.
Specifically, 
% for this comparison, 
we consider a set of five algorithms, which can be broadly classified into three categories, for learning the surrogate model $S$:

\begin{itemize}[noitemsep,nolistsep]
\item \textbf{Directly use the public dataset $\D_{pub}$.} We present two methods to train $S$: 
$\bullet$ \textbf{Direct I.} We train $S$ using the public dataset labelled with target model, {\em i.e.} 
$ \tilde{\D}_{pub}$ with samples  $(x_p, \tilde{y})$, 
$x_p \in \mathcal{D}_{pub}$, 
$\tilde{y} = T(x_p)$;
see Sec.
\ref{ssec:baselineForSurrogate}.
% $\{(x_p,\tilde{y})\}$ 
% where $\tilde{y} = T(x_p), \D_{pub}=\{x_p\}$.
%$ \tilde{y} = T(x), \D_{pub}=\{x_p\}$.  
$\bullet$ \textbf{Direct II.} We apply data augmentation to $ \tilde{\D}_{pub}$ of Direct I to reduce the class imbalance in Direct I, followed by training $S$ using the newly more balanced dataset.
% $\bullet$ \textbf{(A.3)} We apply the SOTA model stealing using hard label only to train the surrogate model $S$. Specially, we use DFMS-HL \cite{sanyal2022towards}. Further details of applying DFMS-HL can be found in the Supplementary. 
\end{itemize}

\begin{itemize}
\vspace{-0.25cm}
\item \textbf{Training an ACGAN.}
We provide two versions: 
$\bullet$ \textbf{ACGAN I.} We train an ACGAN model on $ \tilde{\D}_{pub}$ used in Direct I.  
$\bullet$ \textbf{ACGAN II.} We train an ACGAN model on augmented $ \tilde{\D}_{pub}$ used in Direct II. As $C \circ D$ in ACGAN serves as a classifier, we use $C \circ D$ for MI attacks.

\item \textbf{Training proposed T-ACGAN.} We use our proposed method described in Section \ref{sec:proposed_method} to train T-ACGAN.
Similar to ACGAN I and II, we use $C \circ D$ after training T-ACGAN for the attack.
\end{itemize}

\begin{wraptable}{r}{5cm}
\vspace{-0.5cm}
\caption{We 
compare different approaches to train surrogate model 
for MI attacks. We utilize the following settings: $T =$ FaceNet64, $\D_{priv} =$ CelebA, $\D_{pub} =$ CelebA, and employ the KEDMI\cite{chen2021knowledge} for MI attacks. }
% \vspace{-0.2cm}
\begin{adjustbox}{width=5cm,center}
\begin{tabular}{lrc}
\hline
\textbf{Algorithm} & \textbf{Attack acc. $\uparrow$} & \textbf{KNN dt. $\downarrow$} \\ \hline
Direct I & 5.87  $\pm$	1.65 & 1936.12 \\
Direct II & 9.60 $\pm$	2.22 & 1890.16 \\ \hline
ACGAN I & 6.47  $\pm$	2.15 & 1771.26 \\ 
ACGAN II & 7.87 $\pm$	3.10 & 1785.20 \\ \hline
T-ACGAN & 42.07 $\pm$	3.46 & 1473.99  \\ \hline
 % & C.2 & 51.00 $\pm$ 4.08 & 1436.80 \\ \hline
 \end{tabular}
\end{adjustbox}
\label{tab:algorithms}
\vspace{-0.8cm}
\end{wraptable}

For this comparison, we utilize the following settings: $T$ = FaceNet64, $\D_{priv}$ = CelebA, $\D_{pub}$ = CelebA. Both ACGAN and T-ACGAN adopt the SNResnet architecture \cite{miyato2018cgans,miyato2018spectral}. To ensure a fair comparison, we use the same architecture as $C \circ D$ in ACGAN and T-ACGAN for the surrogate model $S$ in Direct I and Direct II.
Detailed architecture specifications can be found in the Supplementary. 
After training the models, we employ the widely-used KEDMI \cite{chen2021knowledge} as the white-box attack on the trained surrogate models.
Table \ref{tab:algorithms} presents the results. 
The effectiveness of T-ACGAN in training  surrogate models for MI attacks can be observed. 

% \subsection{Surrogate model's accuracy}
\subsection{Comparison against SOTA label-only MI attack}
\label{sec:main_results}

\textbf{Standard MI attack setup.} In this section, we present the results obtained from the standard attack setup on three datasets: CelebA, Facescrub, and Pubfig83, as detailed in Table \ref{tab:mainresults}. 
We evaluate three designs of surrogate:
%We employ three surrogate models as follows:
$\bullet$ (i) We directly use $C \circ D$ from our T-ACGAN as the surrogate model. The architecture of T-ACGAN can be found in the supplementary material.
$\bullet$ (ii) We utilize the synthetic data generated by $G$ of our T-ACGAN and label it using the target classifier $T$ to train another surrogate model, denoted as $S = $ Densenet-161 \cite{huang2017densely}.
$\bullet$ (iii) We employ the same data as in (ii) to train an ensemble of surrogate models, denoted as $S_{en}$, using different architectures including Densenet-121, Densenet-161, and Densenet-169.

We compare our results with the state-of-the-art (SOTA) label-only MI attack BREPMI \cite{kahla2022label}. To conduct our attacks, we utilize white-box PLGMI \cite{yuan2023pseudo} on the surrogate models. Since PLGMI performs attacks using a conditional GAN trained with 
white-box access of 
the target classifier, we replace it with our T-ACGAN, which becomes available for use after training the surrogate models.

Our proposed method LOKT demonstrates a significant improvement in Attack accuracy and KNN distance compared to the SOTA label-only MI attack BREPMI \cite{kahla2022label}. Our top 1 attack accuracies are better than BREPMI from from 17.2\% to 29.87\% across all setups when we utilize the ensemble $S_{en}$.

Fig. \ref{fig:framework} (f) presents a visual comparison of various methods under the setup $\D_{priv} =$ CelebA, $\D_{pub}$ = CelebA. \textbf{More results are available in the Supplementary}. Results clearly indicate that LOKT produces images that are closer to the ground truth (private data) compared to BREPMI \cite{kahla2022label}. This outcome provides strong evidence of the effectiveness of our approach in generating realistic images that closely resemble private data, which is critical for conducting successful MI attacks.

\begin{table}[t]
 \setlength{\tabcolsep}{2.5pt}
\caption{We conduct comprehensive 
experiments to compare our proposed method LOKT 
and existing SOTA BREPMI \cite{kahla2022label} across standard MI attack benchmarks. 
Specifically, we evaluate the performance of our three proposed  designs 
of surrogate, 
% method by employing three surrogate models, 
namely $C \circ D$, $S$, and $S_{en}$, while BREPMI performs black-box search on $T$ directly.
%utilizes $T$ to carry out the attacks. 
We highlight the best results in each setup in \textbf{bold}.
}

\begin{adjustbox}{width=1.0\columnwidth,center}
\begin{tabular}{llllcllllc}
\cmidrule(lr){1-5}\cmidrule(lr){6-10}
\multicolumn{1}{c}{\textbf{Setup}} & \multicolumn{2}{c}{\textbf{Attack}} & \multicolumn{1}{c}{\textbf{Attack acc. $\uparrow$}} & \multicolumn{1}{c}{\textbf{KNN dt. $\downarrow$}} & \multicolumn{1}{c}{\textbf{Setup}} & \multicolumn{2}{c}{\textbf{Attack}} & \multicolumn{1}{c}{\textbf{Attack acc. $\uparrow$}} & \multicolumn{1}{c}{\textbf{KNN dt. $\downarrow$}} \\ 
\cmidrule(lr){1-5}\cmidrule(lr){6-10}

\multirow{4}{2.8cm}{$T$ \hspace{0.48cm} = FaceNet64  $\D_{priv}$ = CelebA  $\D_{pub}$ \hspace{0.01cm} = CelebA}   & \multicolumn{2}{l}{BREPMI} & 73.93   $\pm $ 4.98 & 1284.41 &
\multirow{4}{2.8cm}{$T$ \hspace{0.48cm} = FaceNet64 $\D_{priv}$ = Pubfig83 $\D_{pub}$ \hspace{0.01cm} = Pubfig83 }  & \multicolumn{2}{l}{BREPMI} & 55.60   $\pm $ 4.34 & 1012.83 \\ \cmidrule(lr){2-5}\cmidrule(lr){7-10}
 & \multirow{3}{*}{\textbf{LOKT}}  & $C   \circ D$  & 81.00 $\pm $   4.79 & 1298.63 &  & \multirow{3}{*}{\textbf{LOKT}}  & $C   \circ D$  & 74.80 $\pm $   5.93 & 924.58 \\
 & & $S$  & 92.80 $\pm $   2.59 & 1207.25 &  & & $S$  & 61.60	$\pm $ 3.58 &	993.44  \\
 & & $S_{en}$  & \textbf{93.93 $\pm $   2.78} & \textbf{1181.72} &  & & $S_{en}$  & \textbf{80.00 $\pm $  3.16} & \textbf{883.52} \\  
\cmidrule(lr){1-5}\cmidrule(lr){6-10}
 
\multirow{4}{2.8cm}{$T$ \hspace{0.48cm} = IR152  $\D_{priv}$ = CelebA  $\D_{pub}$ \hspace{0.01cm} = CelebA}  & \multicolumn{2}{l}{BREPMI} & 71.47   $\pm $ 5.32 & 1277.23 & 
\multirow{4}{2.8cm}{$T$ \hspace{0.48cm} = FaceNet64 $\D_{priv}$ = Pubfig83 $\D_{pub}$ \hspace{0.01cm} = FFHQ}  & \multicolumn{2}{l}{BREPMI} & 72.80   $\pm $ 3.90 & 971.51 \\ \cmidrule(lr){2-5}\cmidrule(lr){7-10}
 & \multirow{3}{*}{\textbf{LOKT}}  & $C   \circ D$  & 72.07 $\pm $   4.03 & 1358.94 &  & \multirow{3}{*}{\textbf{LOKT}}  & $C   \circ D$  & 85.60 $\pm $   2.61 & 914.15 \\
 & & $S$  & 89.80 $\pm $   2.33 & 1220.00 &  & & $S$  & 88.40 $\pm $   2.97 & 920.99 \\
 & & $S_{en}$  & \textbf{92.13 $\pm $   2.06} & \textbf{1206.78} &  & & $S_{en}$  & \textbf{94.40 $\pm $   3.85} & \textbf{862.24} \\ 
\cmidrule(lr){1-5}\cmidrule(lr){6-10}

\multirow{4}{2.8cm}{$T$ \hspace{0.48cm} = VGG16  $\D_{priv}$ = CelebA  $\D_{pub}$ \hspace{0.01cm} = CelebA}  & \multicolumn{2}{l}{BREPMI} & 57.40   $\pm $ 4.92 & 1376.94 & 
\multirow{4}{2.8cm}{$T$ \hspace{0.48cm} = FaceNet64 $\D_{priv}$ = Facescrub $\D_{pub}$ \hspace{0.01cm} = Facescrub }  & \multicolumn{2}{l}{BREPMI} & 40.20   $\pm $ 6.60 & 1236.4 \\ \cmidrule(lr){2-5}\cmidrule(lr){7-10}
 & \multirow{3}{*}{\textbf{LOKT}}  & $C   \circ D$  & 71.33 $\pm $   4.39 & 1364.47 &  & \multirow{3}{*}{\textbf{LOKT}}  & $C   \circ D$  & 45.70 $\pm $   4.00 & 1296.29 \\
 & & $S$  & 85.60 $\pm $   3.03 & 1252.09 &  & & $S$  & 53.20 $\pm $   5.29 & 1280.70 \\
 & & $S_{en}$  & \textbf{87.27 $\pm $   1.97} & \textbf{1246.71} &  & & $S_{en}$  & \textbf{58.60 $\pm $   4.86} & \textbf{1225.13} \\  
\cmidrule(lr){1-5}\cmidrule(lr){6-10}
\multirow{4}{2.8cm}{$T$ \hspace{0.48cm} = FaceNet64  $\D_{priv}$ = CelebA  $\D_{pub}$ \hspace{0.01cm} = FFHQ}   & \multicolumn{2}{l}{BREPMI} & 43.00   $\pm $ 5.14 & 1470.55 & 
\multirow{4}{2.8cm}{$T$ \hspace{0.48cm} = FaceNet64 $\D_{priv}$ = Facescrub $\D_{pub}$ \hspace{0.01cm} = FFHQ} & \multicolumn{2}{l}{BREPMI} & 37.30   $\pm $ 3.99 & 1456.59 \\ \cmidrule(lr){2-5}\cmidrule(lr){7-10}
 & \multirow{3}{*}{\textbf{LOKT}}  & $C   \circ D$  & 43.27 $\pm $   3.53 & 1516.18 &  & \multirow{3}{*}{\textbf{LOKT}}  & $C   \circ D$  & 44.50 $\pm $   5.98 & 1403.73 \\
 & & $S$  & 59.13 $\pm $   2.77 & 1437.86 &  & & $S$  & 47.20 $\pm $   4.39 & 1404.85 \\
 & & $S_{en}$  & \textbf{62.07 $\pm $   3.89} & \textbf{1428.04} &  & & $S_{en}$  & \textbf{53.70 $\pm $   4.57} & \textbf{1338.67} \\  
\cmidrule(lr){1-5}\cmidrule(lr){6-10}
\end{tabular}
\end{adjustbox}
\label{tab:mainresults}
\vspace{-0.5cm}
\end{table}

\textbf{MI attacks under large distribution shift.}
Table \ref{tab:mainresults} compares the MI attack results in the large distribution shift setup, where we use $\D_{pub}$ = FFHQ, $\D_{priv}$ = CelebA/ Facescrub/ Pubfig83, and $T = $ FaceNet64. The attack results of BREPMI drop significantly (by 30.93\% for CelebA and 2.9\% for Facescrub), while the results for Pubfig83 notably increase, which can be attributed to the small size of the Pubfig83 dataset \cite{kahla2022label}. Our proposed method outperforms BREPMI, with the top 1 attack accuracies increase from 16.40\% to 21.60\% for all setups.
% with better top 1 attack accuracy ranging from 16.4\% to 21.6\%. 
Moreover, the KNN distance indicates that our reconstructed images are closer to the private data than those reconstructed by BREPMI. 

% Remarkably, with access only to the hard label of the target classifier $T$, our approach outperforms two white-box attacks, GMI and KEDMI, where the adversary has full access to the target classifier $T$. Our results also demonstrate that our approach closes the gap with the SOTA white-box MI attack, PLGMI, which is most currently white-box attack obtaining a notable attack results. 

\subsection{Additional results}
\label{sec:additional_results}

% In summary, our experiments showcase the robustness and versatility of our surrogate models, thereby underlining their potential for use in diverse MI applications.

\textbf{MI attack results using MI defense model.}
We investigate the attacks on the  MI defense model (see Table \ref{tab:bido}). 
% Currently, defense models focus on defending against white-box attacks as black-box and label-only MI attacks are not fully noticeable. 
Specifically, we utilize the SOTA defense model BiDO-HSIC \cite{peng2022bilateral} and MID \cite{wang2021improving}. 
Our results indicate that BiDO-HSIC successfully reduce the effectiveness of the white-box SOTA attack, PLGMI, by 9.57\% (See the result in the Supplementary).  
In the label-only setup, the performance of BREPMI becomes relatively low with attack accuracy of only 37.40\% for BiDO-HSIC \cite{peng2022bilateral} and 39.20\% for MID \cite{wang2021improving}. In contrast, our approach achieves a much higher attack accuracy of 60.73\% and 60.33\%, almost doubling the performance of BREPMI. These results demonstrate that our approach is effective in conducting MI attacks on MI defense models, even in scenarios where the adversary has limited information about the target classifier.

\begin{table}[t]%!htb]

\begin{minipage}{.48\linewidth}

\caption{ 
We report Label-only MI Attack results under SOTA defense models namely BiDO \cite{peng2022bilateral} and MID \cite{wang2021improving}.
% using label-only attack BREPMI \cite{kahla2022label} and our proposed method LOKT. 
We use $\D_{priv} =$ CelebA, $\D_{pub} =$ CelebA. We highlight the best results in \textbf{bold}.
}

 \setlength{\tabcolsep}{2.5pt}
\begin{adjustbox}{width=1.0\columnwidth,center}
\label{tab:bido}
\begin{tabular}{llllc}
\hline
\multicolumn{1}{c}{\textbf{Setup}} & \multicolumn{2}{c}{\textbf{Attack}} & \multicolumn{1}{c}{\textbf{Attack acc. $\uparrow$}} & \multicolumn{1}{c}{\textbf{KNN dt. $\downarrow$}} \\ \hline
\multirow{4}{2.5cm}{$T$ = BiDO \cite{peng2022bilateral} $\D_{priv}$ = CelebA  $\D_{pub}$ \hspace{0.01cm} = CelebA}   &  \multicolumn{2}{c}{BREPMI\cite{kahla2022label}} & 37.40 $\pm$ 3.66  & 1500.45 \\ \cmidrule(lr){2-5}
& \multirow{3}{*}{\textbf{LOKT}}  & $C   \circ D$  & 45.73 $\pm$ 5.94 & 1493.48 \\
& & $S$ & 58.53 $\pm$ 4.87 & 1427.22\\
& & $S_{en}$ &  \textbf{60.73 $\pm$ 3.07} & \textbf{1395.93} \\ \hline
\multirow{4}{2.5cm}{$T$ \hspace{0.48cm} = MID \cite{wang2021improving}  $\D_{priv}$ = CelebA  $\D_{pub}$ \hspace{0.01cm} = CelebA}   & \multicolumn{2}{c}{BREPMI\cite{kahla2022label}} & 39.20	$\pm$ 4.19 & 1458.61 \\ \cmidrule(lr){2-5}
 & \multirow{3}{*}{\textbf{LOKT}} & $C \circ D$ & 44.13 $\pm$ 3.54	& 1475.73 \\
& & $S$ & 55.33 $\pm$ 4.40 & 1393.76 \\
& & $S_{en}$ & \textbf{60.33 $\pm$ 
 4.76} & \textbf{1374.34} \\
\hline
\end{tabular}
\end{adjustbox}

\end{minipage}%
\hspace{0.2cm}
\begin{minipage}{.48\linewidth}
\centering
\caption{The comparison of the number of queries (in millions) used by LOKT and BREPMI \cite{kahla2022label}. 
% Here $C \circ D$ and $S =$ Densenet-161, $S_{en} = \{$Densenet-121, Densenet-161, Densenet-169$\}$. 
The attacks using $S$ and $S_{en}$ consume additional 500k queries comparing to $C \circ D$ to label the synthetic images to train $S$ and $S_{en}$. Our results show that we use fewer number of queries than BREPMI in all setups.}
\label{tb:query_budget}
\begin{adjustbox}{width=0.98\columnwidth,center}
\begin{tabular}{lccc} 
\hline
\textbf{$T$} & \textbf{LOKT $C \circ D$} & \textbf{LOKT $S/S_{en}$} & \textbf{BREPMI} \\ \hline
FaceNet64 & 12.16 & 12.66 & 17.98 \\
IR152 & 12.16 &12.66 & 18.06 \\
VGG16 & 12.16 &12.66 & 18.12  \\
BiDO-HSIC & 12.16 & 12.66 & 18.39  \\
MID & 12.16 & 12.66 & 18.25 \\ \hline 
\end{tabular}
\end{adjustbox}
\end{minipage} 
\vspace{-0.5cm}
\end{table}

\textbf{High resolution.} We conduct the experiment with high resolution images which has not been addressed yet for label-only setup \cite{kahla2022label}. 
In particular, we train a new target classifier $T$ = Resnet-152 using CelebA setup with the image size = 128$\times$128. For fair comparison between BREPMI and our proposed method, T-ACGAN has the same GAN architectures used by BREPMI. The details of the architecture can be found in the Supplementary.

The results are shown in Table \ref{tab:highres_128}. 
LOKT outperforms BREPMI, with top 1 accuracy surpassing BREPMI by 20.27\%. Our inverted images are closer to private training samples than BREPMI (smaller KNN distance). We believe our study can provide new insight on the effectiveness of SOTA label-only attack at a higher resolution of 128$\times$128, paving the way to future label-only model inversion attacks at resolutions beyond 128$\times$128.

\textbf{Query budget.} 
In this experiment, we compare  query budget between our proposed method and BREPMI \cite{chen2021knowledge}. 
In the BREPMI, queries to the target classifier $T$ are required to  identify the initial points for attacking and estimate the gradients during the attack. In our  method, 
queries to  $T$ are required to label the synthetic data during the training of
T-ACGAN to obtain $C \circ D$, 
and additional 500k queries to label generated images of T-ACGAN to train $S$ and the ensemble $S_{en}$.
For comparison, as shown in Table \ref{tb:query_budget}, we use $\D_{priv} =$ CelebA and $\D_{pub} =$ CelebA.
The results show that our proposed method requires 
30\% fewer queries compared to BREPMI.

\subsection{User study}
\label{main-sub-sec:user-studies}
\textbf{User study setup.} 
In this section, we go beyond objective metrics and consider subjective evaluation of MI attacks.
In particular, we conduct a human study to understand the efficacy of our proposed method, LOKT, compared to BREPMI.
We follow the setup by \cite{an2022mirror} for human study and use Amazon Mechanical Turk (MTurk) for experiments. 
The user interface is provided in the Supplementary.
In this study, users are shown 5 real images of a person (identity) as reference. Then users are required to compare the 5 real images with two inverted images: one from our method (LOKT), the other from BREPMI. We use $D_{priv}$  = CelebA, $D_{pub}$ = CelebA and $T$  = FaceNet64. Following \cite{an2022mirror}, we randomly selected 50 identities with 10 unique users evaluating each task accounting to 1000 comparison pairs.

\textbf{User study results.} 
We report the user study results in Table \ref{tab:study1}. Our  human study reveals that users distinctly favor our approach, with 64.30\% user preference for images reconstructed using our proposed approach, in contrast to BREPMI’s lower 35.70\% user preference. 
These subjective evaluations further show the efficacy of our proposed method, LOKT, in the challenging label-only MI setup. 
% We conduct an additional human study and details are provided in Supplementary.

\begin{table}[t]%!htb]

\begin{minipage}{.5\linewidth}
\centering
\caption{We conduct the experiment with higher resolution images. We use $T$ = Resnet-152, $\D_{priv}$ = CelebA, $\D_{pub}$ = CelebA, image size = 128$\times$128. The natural accuracy of $T$ is 86.07\%. We highlight the best results in \textbf{bold}.
}
\label{tab:highres_128}
\begin{adjustbox}{width=1.0\columnwidth,center}
\begin{tabular}{llllc}
\hline
\multicolumn{1}{c}{\textbf{Setup}} & \multicolumn{2}{c}{\textbf{Attack}} & \multicolumn{1}{c}{\textbf{Attack acc. $\uparrow$}} & \multicolumn{1}{c}{\textbf{KNN dt. $\downarrow$}} \\ \hline
\multirow{4}{2.5cm}{$T$ = IR152 \\$\D_{priv}$ = CelebA  $\D_{pub}$ \hspace{0.01cm} = CelebA}  &  \multicolumn{2}{c}{BREPMI\cite{kahla2022label}} & 50.33 $\pm$ 4.71 & 1389.09 \\ \cmidrule(lr){2-5}
& \multirow{3}{*}{\textbf{LOKT}}  & $C   \circ D$  & 66.87 $\pm$ 3.93  & 1356.53 \\
& & $S$ & 66.80 $\pm$ 3.83 & 1341.04 \\
& & $S_{en}$ &  \textbf{70.60 $\pm$ 4.43} & \textbf{1320.16} \\ \hline
\end{tabular}
\end{adjustbox}
\end{minipage} 
\hspace{0.1cm}
\begin{minipage}{.47\linewidth}

\caption{User study results. Our human study reveals that users distinctly favor our approach, with 64.30\% user preference for images reconstructed using our proposed approach, LOKT, compared to BREPMI’s lower 35.70\% user preference.}
\setlength{\tabcolsep}{2.5pt}
\label{tab:study1}
\begin{adjustbox}{width=0.65\columnwidth,center}
\begin{tabular}{lc} \hline
         \textbf{Method}& \textbf{User Preference ($\uparrow$)}  \\ \hline
         BREPMI & 35.70\%  \\
         LOKT & \textbf{64.30}\% \\ \hline
    \end{tabular}
\end{adjustbox}
\end{minipage}
\vspace{-0.52cm}
\end{table}

\begin{minipage}{.48\linewidth}

\end{minipage}

\begin{minipage}{.48\linewidth}

\end{minipage}
\vspace{-1.2cm}
\section{Discussion}

\textbf{Conclusion.}
Instead of performing a black-box search approach as in existing SOTA, we propose a new label-only MI approach (LOKT) by transferring decision knowledge from the target model to 
surrogate models and performing white-box attacks on the surrogate models.
To obtain the surrogate models, we propose a new T-ACGAN to leverage generative modeling and the target model for effective
knowledge transfer.
Using findings of general learning dynamics of DNNs, 
we conduct  analysis to support that our surrogate models are effective proxies for the target model under MI.
We perform extensive experiments and ablation to support our claims and demonstrate significant improvement over existing SOTA.

\textbf{Broader Impacts.}
% Our  research of understanding model inversion attack is important when AI models are increasingly deployed in many applications.
Understanding model inversion attacks holds significance as AI models continue to see widespread deployment across various applications. By studying and understanding the approaches and methodologies for model inversion, researchers can develop  
good practices in deploying AI models and robust defense mechanisms for different applications esp. those involving sensitive training data.
It is important to emphasize that the objective of model inversion research is to raise  awareness of potential privacy threats and bolster our collective defenses.
%Responsible use of the knowledge gained from this research plays an important role in ensuring that advancements are utilized to understand and maintain  trust in AI models.

{\bf Limitations.} 
% Compared to white-box/black-box MI, label-only MI is the most restrictive setting and is very challenging. The study of label-only attacks on high resolution suggest that both BREPMI \cite{kahla2022label} and our proposed method face significant challenges in reconstructing private training data under high-resolution setup. 
% We believe our study can provide new insight on the effectiveness of SOTA label-only attacks on 128x128, paving the way to future label-only model inversion attacks at higher resolutions.
% While our experiments are extensive compared to previous works, in practical applications, there are other types of private training datasets involved.
% However, as our assumption is general, we believe our findings can generalize to other applications. 
While our experiments are extensive compared to previous works, practical applications involve different types of private training datasets such as healthcare data.
Nevertheless, our assumptions are general, and we believe our findings can be applied to a broader range of applications.

\newpage
\textbf{Acknowledgements.}
This research is supported by the National Research Foundation, Singapore under its AI Singapore Programmes
(AISG Award No.: AISG2-TC-2022-007) and SUTD project PIE-SGP-AI-2018-01. 
This research work is also supported by the Agency for Science, Technology and Research (A*STAR) under its MTC Programmatic Funds (Grant No. M23L7b0021). 
This material is based on the research/work support in part by the Changi General Hospital and Singapore University of Technology and Design, under the HealthTech Innovation Fund (HTIF Award No. CGH-SUTD-2021-004).
We thank anonymous reviewers for their insightful feedback.
% % the National Research Foundation, Singapore under its AI Singapore Programmes (AISG Award No.: AISG2-RP-2021-021; AISG Award No.: AISG2-TC-2022-007). 
% % This project is also supported by SUTD project PIE-SGP-AI-2018-01. 

{\small
\bibliographystyle{unsrt}
% \bibliography{main}
\bibliography{egbib}%2.bbl}
}

\newpage
\appendix
\unhidefromtoc

{\centering \large \textbf{Supplementary Materials}}

In this supplementary material, we provide additional experiments, analysis, ablation study, and details required to reproduce our results. Pytorch code, demo, pre-trained models and reconstructed data are available at our 
% \textcolor{magenta} {\href{https://ngoc-nguyen-0.github.io/label_only_tacgan/}{project website}}.
\textcolor{magenta} {\href{https://ngoc-nguyen-0.github.io/lokt/}{project website}}.

\thispagestyle{empty}

\tableofcontents
\clearpage
\renewcommand\thefigure{\thesection.\arabic{figure}}
\renewcommand\theHfigure{\thesection.\arabic{figure}}
\renewcommand\thetable{\thesection.\arabic{table}}  
\renewcommand\theHtable{\thesection.\arabic{table}}  

\setcounter{figure}{0} 
\setcounter{table}{0} 

% Keshik: Analysis (Add experiment details)
\section{Additional Analysis and Visualizations}
\subsection{Our Surrogate models are effective proxies for the opaque Target model for MI}

% Part 1
%%% =============== Main Figure for training set =============
\begin{figure}[!h]
\begin{adjustbox}{width=1.0\textwidth,center}
\begin{tabular}{c}
    \includegraphics[width=0.99\textwidth]{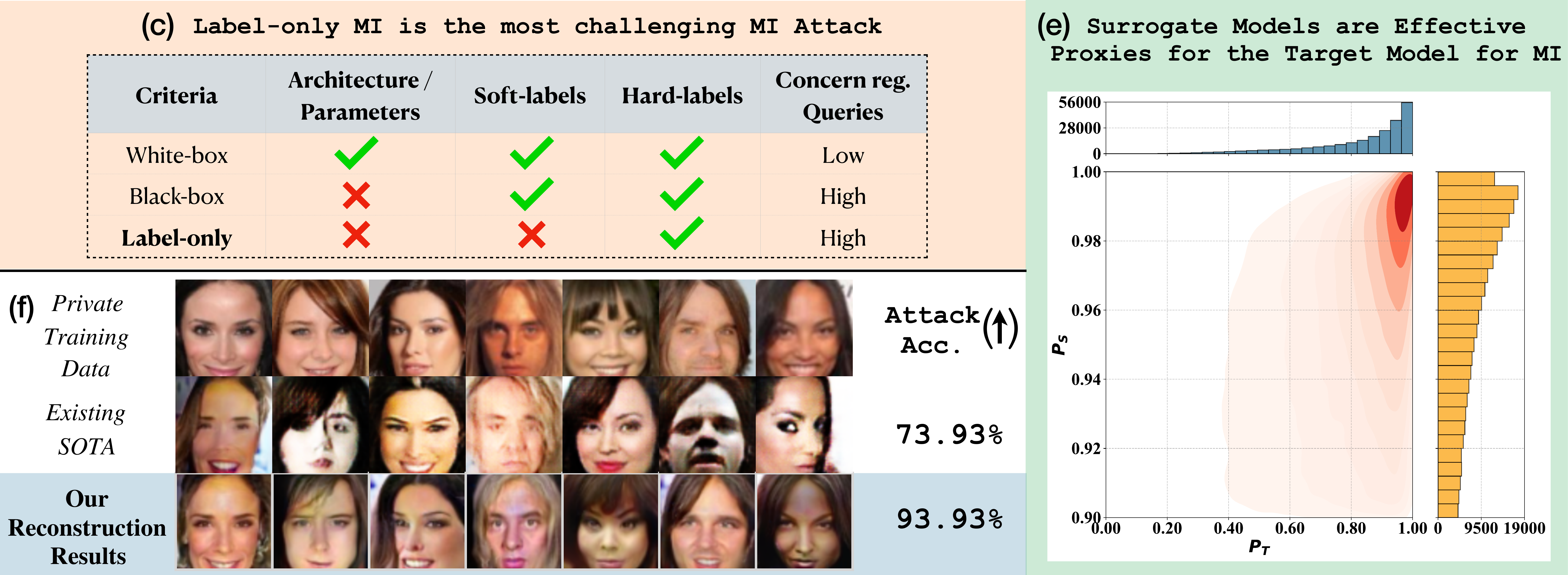}
\end{tabular}
\end{adjustbox}
% \vspace{-0.35cm}
\caption{
\small
We use $\D_{priv}$ = CelebA, $\D_{pub}$ = CelebA, $T$ = FaceNet64, $S$ = DenseNet-161.
{\bf (c)} We cast the challenging problem setup of label-only MI attack as a white-box MI attack. To our knowledge, our proposed approach is the first to address label-only MI via white-box MI attacks.
{\bf (e)} We consider high likelihood samples under $S$. i.e.: $P_S>0.9$. 
Our analysis using 500k training data demonstrates that $S$ is an effective proxy for $T$ for MI attack.
In particular, the white-box MI attack on $S$ mimics the white-box attack on opaque $T$. 
{\bf (f)} Additional reconstruction results using our proposed approach ($S_{en}$). We remark that our proposed approach significantly improves the Label-only MI attack (e.g. $\approx$ 20\% improvement in standard CelebA benchmark compared to existing SOTA \cite{kahla2022label}) resulting in significant improvement in private data reconstruction.
Best viewed in color.
}
\label{fig_supp:celeba_celeba_facenet64_analysis}
%\vspace{-0.5cm}
\end{figure}

% \begin{wrapfigure}[19]{r}{0.42\textwidth}
\begin{figure}[ht]
\centering
% \vspace{-0.5cm}
    \includegraphics[width=0.46\textwidth]{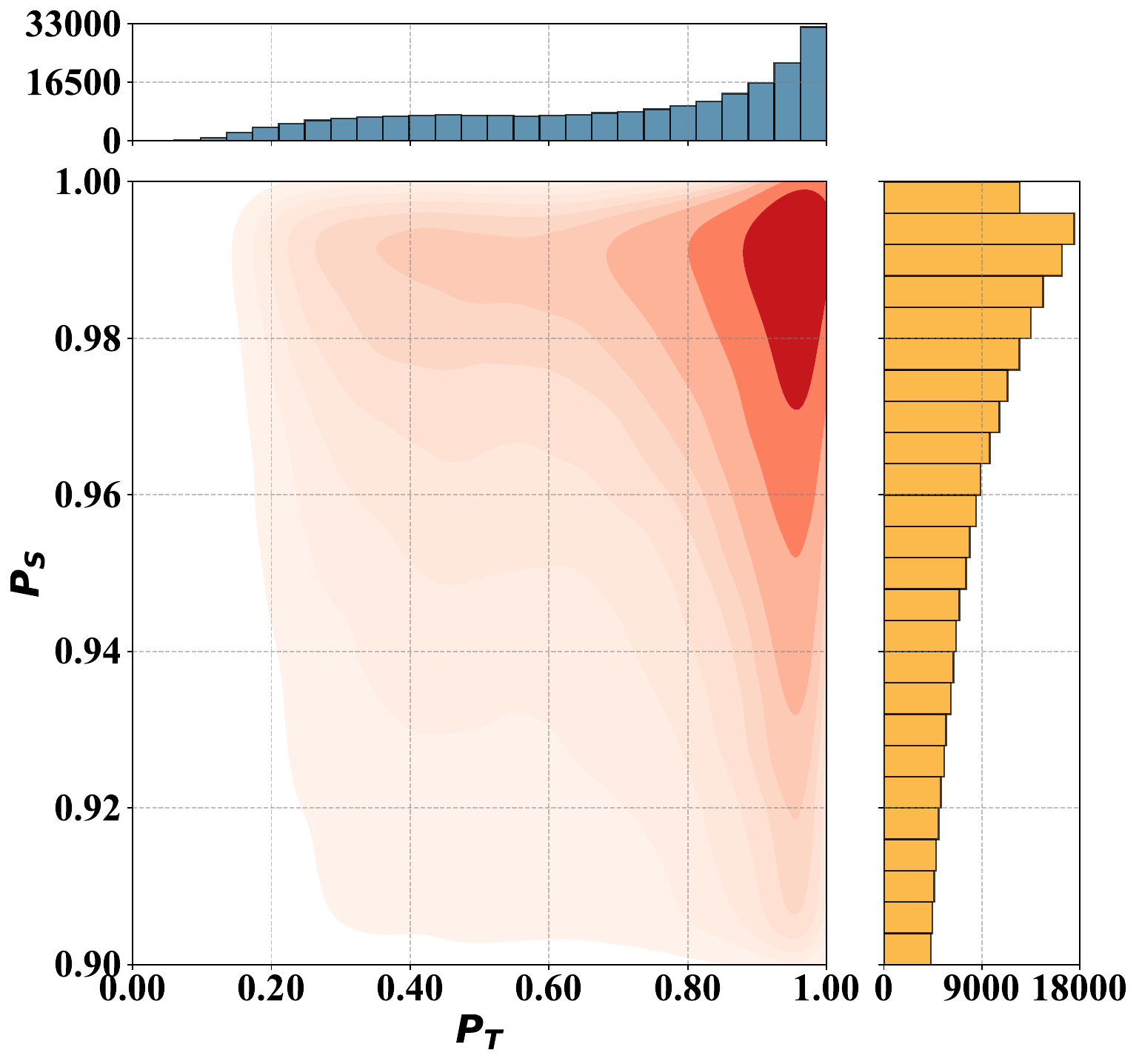}
% \vspace{-0.3cm}
\caption{
% \small
Figure 1 (e) from the main paper supports that $S$ is a good proxy for $T$ for MI established using \textcolor{magenta}{\bf Property P1}.
We use $\D_{priv}$= CelebA, $\D_{pub}$ = CelebA, $T$ = FaceNet64, $S$ = DenseNet-161. 
We use 500k validation data for analysis.
}
\label{fig_supp:fig_1e_main_paper}
\end{figure}
% \end{wrapfigure}

\begin{figure}[!h]
\centering
\begin{tabular}{cc}
    {\includegraphics[width=0.49\linewidth]{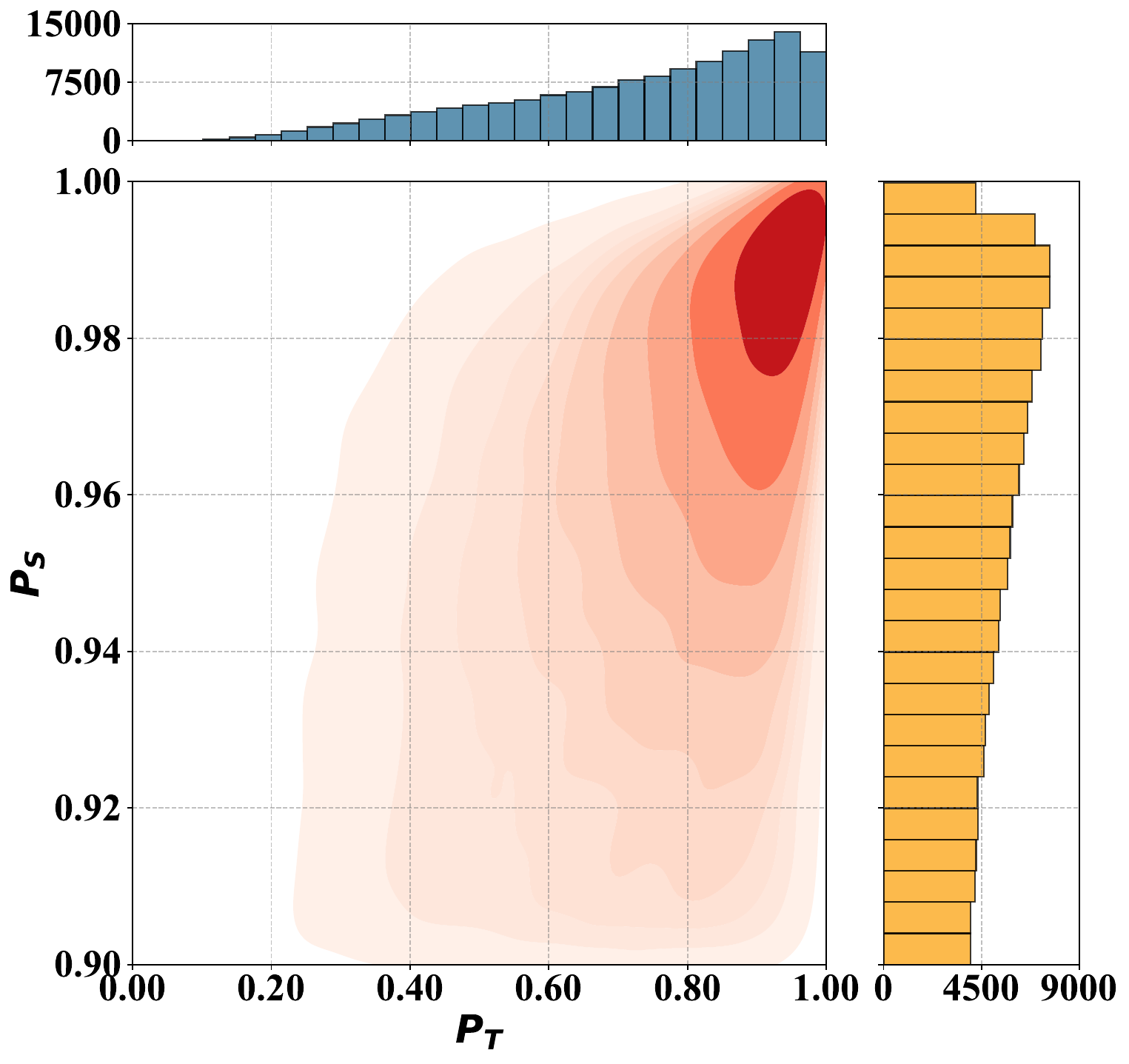}} & 
    {\includegraphics[width=0.49\linewidth]{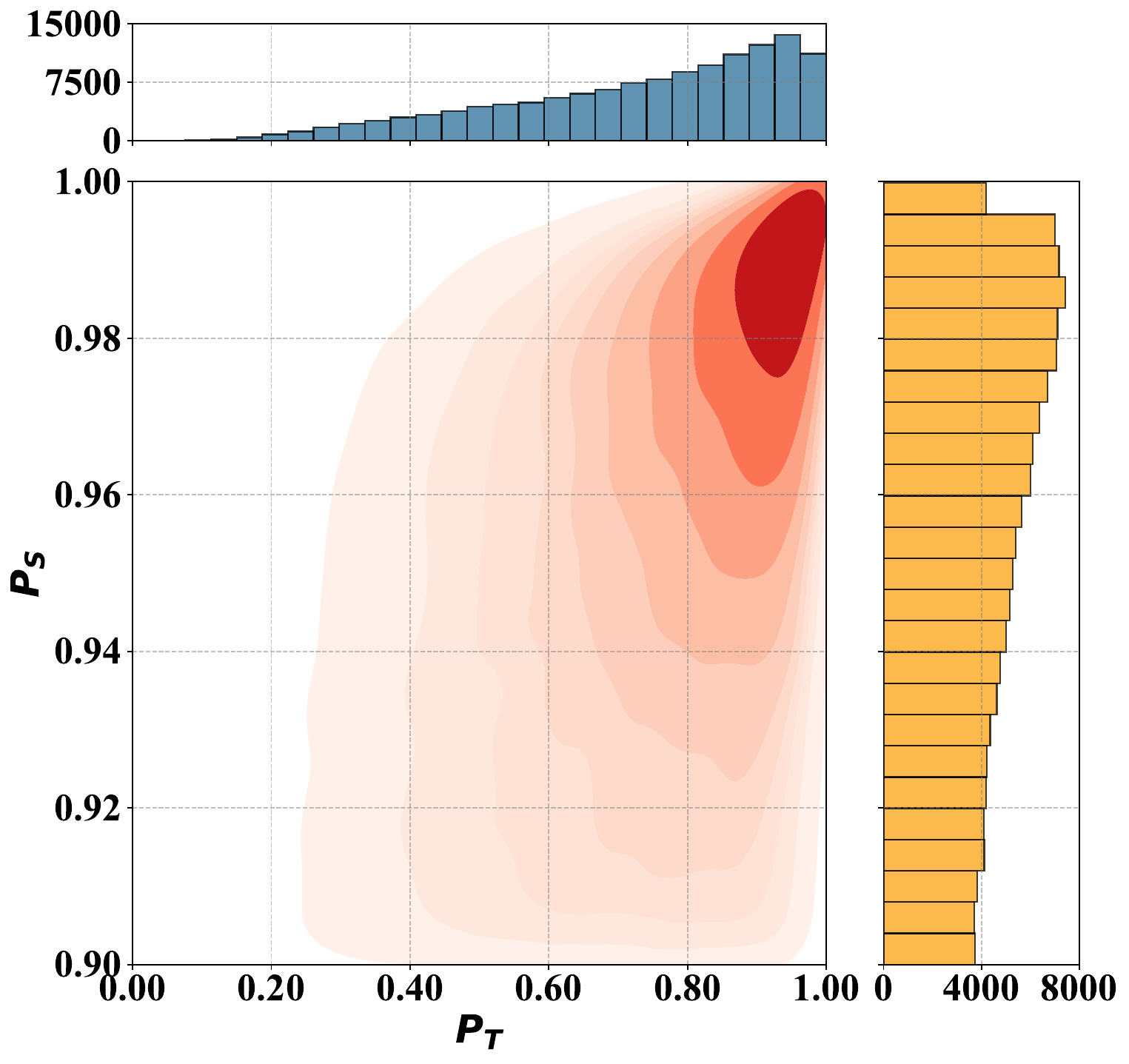}}
    
\end{tabular}
\vspace{-0.3cm}
\caption{
We use $\D_{priv}$ = CelebA, $\D_{pub}$ = FFHQ, $T$ = FaceNet64, $S$ = DenseNet-161.
we consider high likelihood samples under $S$. i.e.: $P_S~>~0.90$, and show results for 500k training samples (left) and 500k validation samples (right).
As one can clearly observe in both conditional $P_T$ histograms,  high likelihood samples under $S$ are likely to have high likelihood under $T$ \textcolor{magenta}{\bf (Property P1)}, and 
it is uncommon for high likelihood samples under $S$ to have low likelihood under $T$.
Given \textcolor{magenta}{\bf P1}, white-box attacks on $S$ can mimic white-box attacks on $T$, resulting in $S$ being a an effective proxy of $T$ for MI.
}
\label{fig_supp:ps_pt_analysis_celeba_ffhq_facenet64}
\vspace{-0.1cm}
\end{figure}

\textbf{White-box MI attack on S mimics the white-box attack on T.} 
For clarity, we copy {\color{red}Figure 1(e)} (main paper) to Supplementary Fig. \ref{fig_supp:fig_1e_main_paper}.
In this section, we include the details of Fig. \ref{fig_supp:fig_1e_main_paper} and provide additional empirical evidence in Figure \ref{fig_supp:celeba_celeba_facenet64_analysis}(e) to support \textcolor{magenta}{\bf Property P1}.
We remark that Fig. \ref{fig_supp:fig_1e_main_paper} and Fig. \ref{fig_supp:celeba_celeba_facenet64_analysis}(e) use 500k validation and 500k training data respectively\footnote{We recall that the data samples are generated samples from our T-ACGAN. Using generated samples for analysis is suitable as generated samples are utilized during white-box MI.}.
% We use $\D_{priv}$=CelebA, $\D_{pub}$=CelebA, $T$=FaceNet64, $S$=DenseNet-161 for analysis.
% We generate 500k synthetic data from our proposed T-ACGAN for training and validation purposes.
In both figures, we consider high-likelihood samples under $S$. i.e.: $P_S~>~0.90$.
We remark that since in our framework, we optimize white-box attack w.r.t. $S$, the reconstructed samples usually have a very high likelihood under $S$ (above 0.9).
%, i.e., $P_S \approx 0.99$. 
%man: why not we condition on 0.99 
Therefore, we condition our analysis on $P_S > 0.9$.
As one can clearly observe in both conditional $P_T$ histograms in Fig. 
\ref{fig_supp:fig_1e_main_paper} and Fig. \ref{fig_supp:celeba_celeba_facenet64_analysis}(e), high likelihood samples under $S$ are likely to have high likelihood under $T$ \textcolor{magenta}{\bf (Property P1)}, and it is uncommon for high likelihood samples under $S$ to have low likelihood under $T$.
Given \textcolor{magenta}{\bf P1}, white-box attacks on $S$ can mimic white-box attacks on $T$, resulting in $S$ being an effective proxy for $T$ for MI.
In addition, we report similar observations on another setup: $\D_{priv}$=CelebA, $\D_{pub}$=FFHQ, $T$=FaceNet64, $S$=DenseNet-161 in Fig. \ref{fig_supp:ps_pt_analysis_celeba_ffhq_facenet64}.

% ---- Main Figure (Show on page 1) ----------

\begin{figure}[!th]
\vspace{-0.5cm}
\begin{adjustbox}{width=1.0\textwidth,center}
\begin{tabular}{c}
    \includegraphics[width=0.99\textwidth]{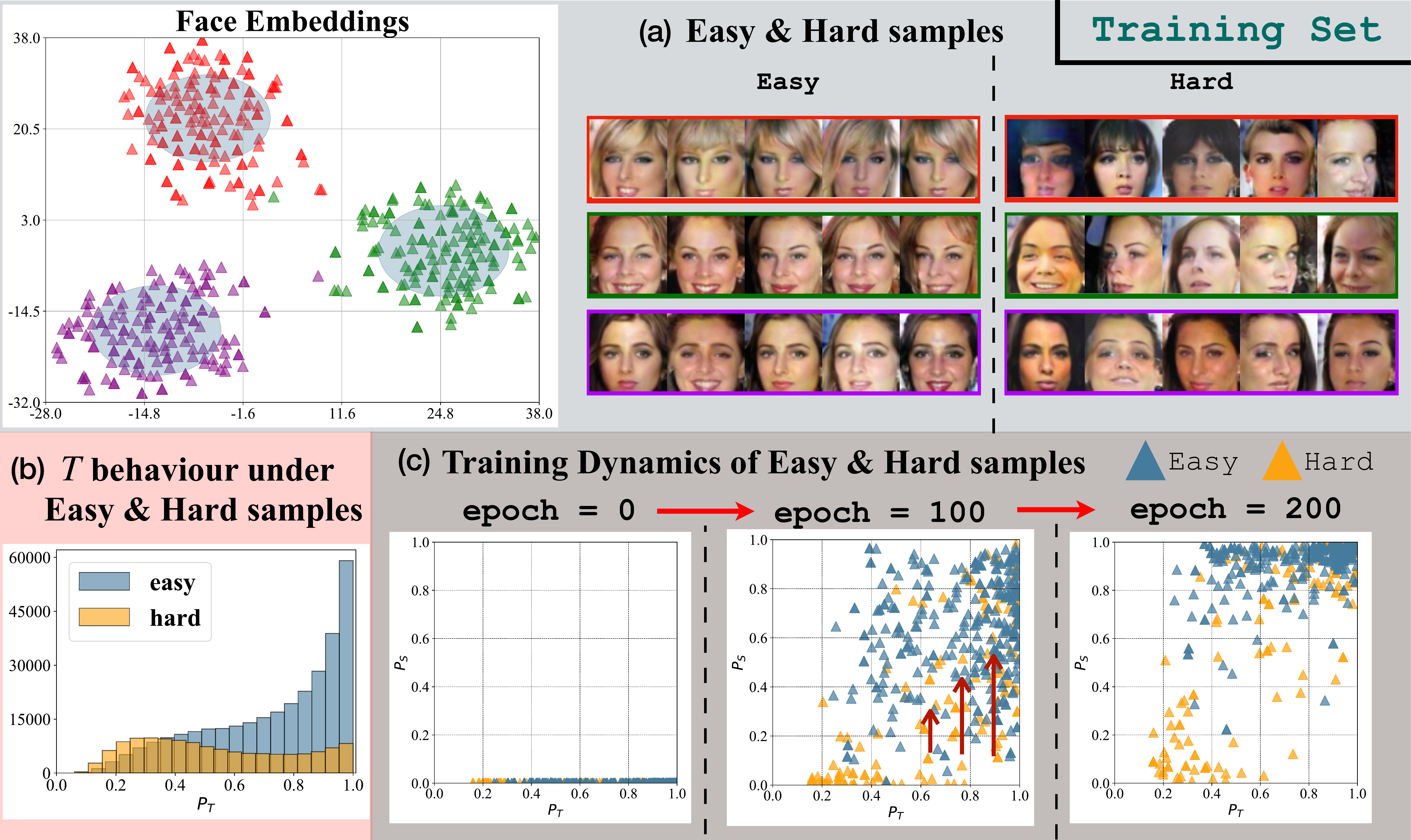}\\
    \hline \\
    \includegraphics[width=0.99\textwidth]{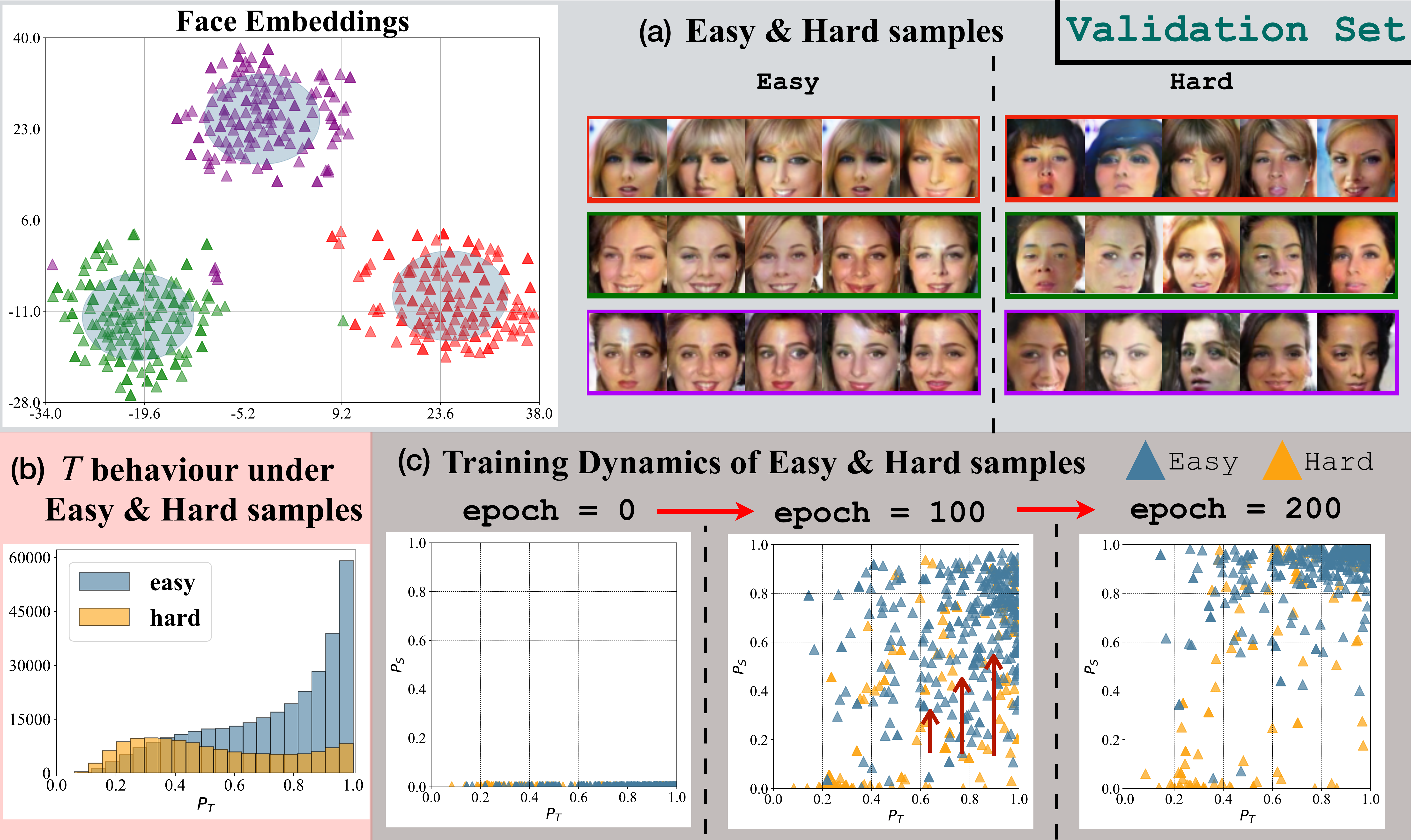}
\end{tabular}
\end{adjustbox}
\vspace{-0.35cm}
\caption{
\small
We use $\D_{priv}$ = CelebA \cite{liu2015deep}, $\D_{pub}$ = CelebA \cite{liu2015deep}, $T$ = FaceNet64, $S$ = DenseNet-161. 
The face embeddings are extracted using publicly available SOTA face recognition models \href{https://github.com/ageitgey/face_recognition}{here}. 
Similar to our main paper, we apply the framework of \cite{arpit2017} to analyze 
learning dynamics of $S$ to reason why $S$ possesses  \textcolor{magenta}{\bf P1},  and therefore could be an effective proxy for $T$ under MI. 
We analyze generated samples $x_f$ from our T-ACGAN for 3 identities (IDs {\color{red} 49}, {\color{ForestGreen} 34}, {\color{Plum} 58}). 
We use 150 samples for each identity, and show results for both training set (top) and validation set (bottom).
Note that $x_f$ analysis is relevant as generated samples are used in MI attacks.
{\bf (a)}: 
Different clusters and different distances from cluster centroids can be observed, suggesting 
patterns of face identities in some samples (easy samples) while
diverse
appearance in 
other samples (hard samples). 
We use distances from centroids to identify easy samples  
$x_f^e$ and hard samples $x_f^h$ (easy samples are indicated using a transparent blue circle for each ID in the visualization). 
Visualization of $x_f^e$ and $x_f^h$ in image space further demonstrates identity patterns in $x_f^e$
and diverse
appearance in $x_f^h$.
{\bf (b)}:
Similar to \cite{arpit2017}, we observe that 
$x_f^e$ and $x_f^h$ tend to have high and low likelihood under $T$ ($P_T$) resp. This is shown using 500k training data (top) and 500k validation data (bottom).
{\bf (c)}:  
We track likelihood under $S$ ($P_S$) for 
$x_f^e$ and $x_f^h$ during the training of $S$.
As training progresses, $P_S$ of $x_f^e$ and $x_f^h$ improve in general, and samples move up vertically (note that $P_T$ of samples do not change).
Consistent with the 
{\em 
``DNNs Learn Patterns First''}
finding
in \cite{arpit2017}, $S$ learns general identity patterns first to fit the easy samples.
Therefore, $P_S$ of $x_f^e$ improve at a faster pace in the training, and many of them achieve high $P_S$
at epoch = 200.
As $x_f^e$ tend to have high $P_T$, we observe property \textcolor{magenta}{\bf P1} in $S$.
% , achieves high likelihood on the easy samples (which also exhibit high likelihood under $T$). Hence we observe property P1 in $S$. 
For $x_f^h$ (many of them tend to have low $P_T$), 
%which exhibit low likelihood under $T$), 
it is uncommon for $S$ to achieve high likelihood on them as they do not fit easily to the pattern learned by $S$.
Best viewed in color.
}
\label{fig_supp:training-dynamics-setup1}
\vspace{-0.5cm}
\end{figure}

\begin{figure}[t]
% \vspace{-0.5cm}
\begin{adjustbox}{width=1.0\textwidth,center}
\begin{tabular}{c}
    \includegraphics[width=0.99\textwidth]{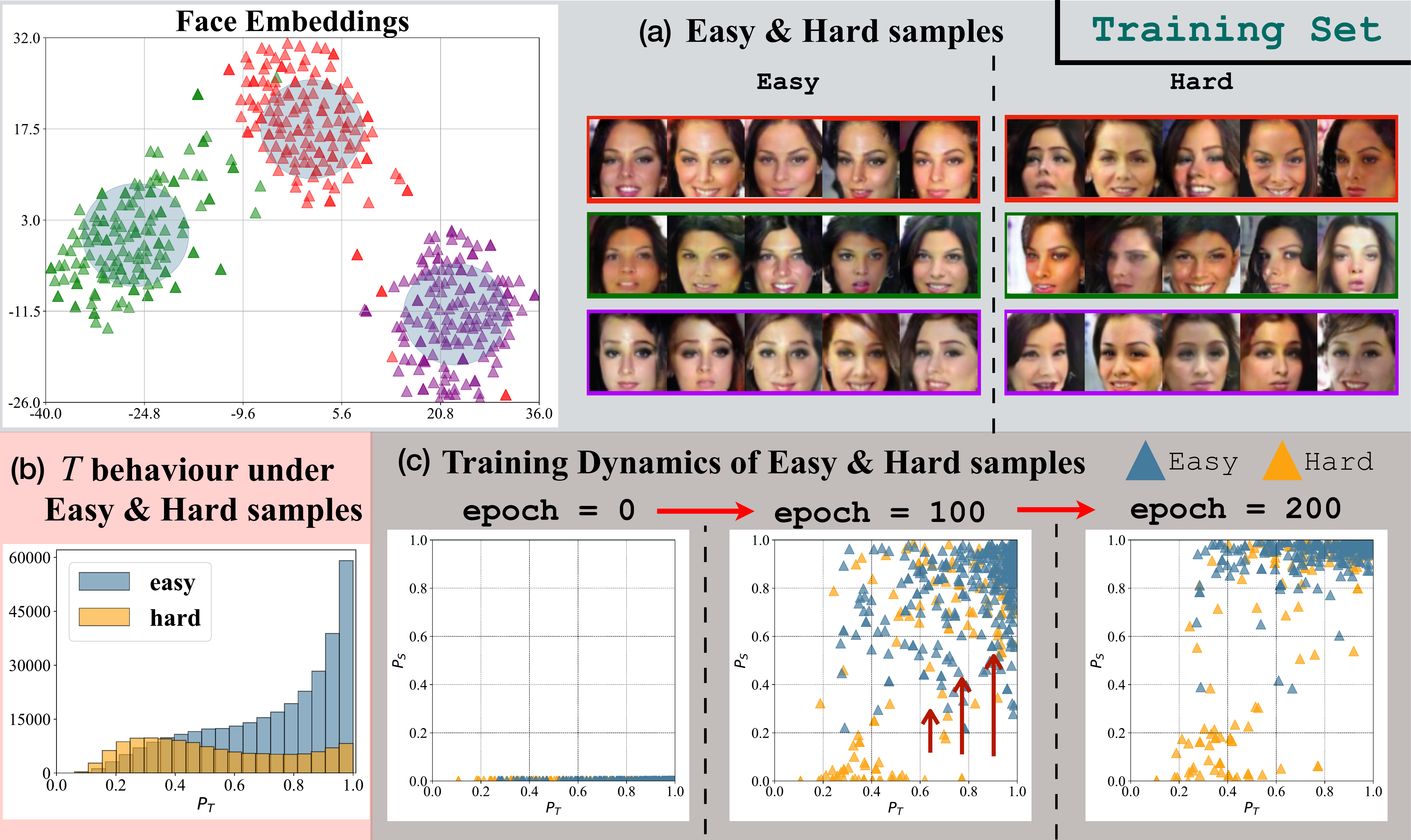} \\
    \hline
    \includegraphics[width=0.99\textwidth]{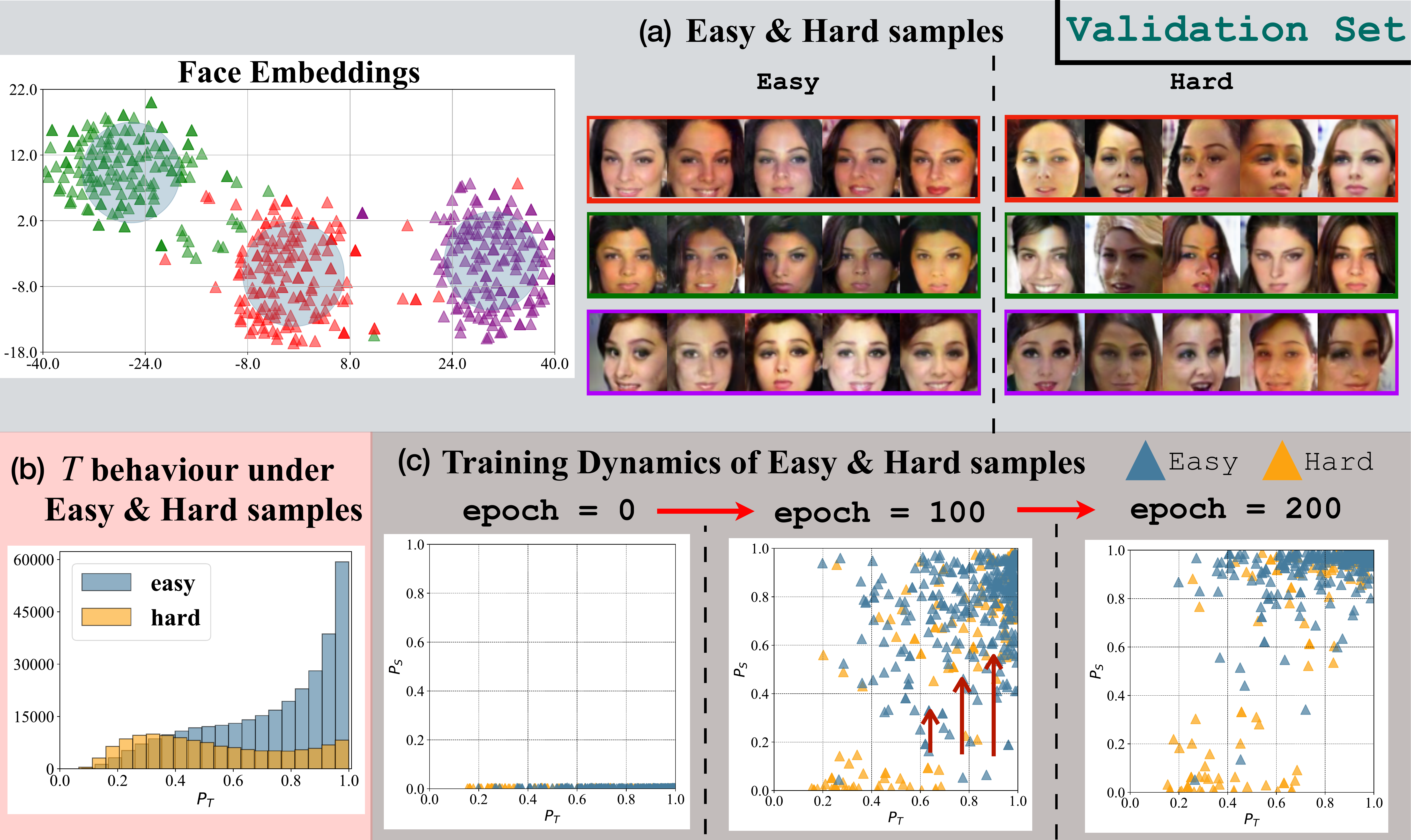}
\end{tabular}
\end{adjustbox}
\vspace{-0.35cm}
\caption{
\small
We use $\D_{priv}$ = CelebA \cite{liu2015deep}, $\D_{pub}$ = CelebA \cite{liu2015deep}, $T$ = FaceNet64, $S$ = DenseNet-161. 
The face embeddings are extracted using publicly available SOTA face recognition models \href{https://github.com/ageitgey/face_recognition}{here}. 
Similar to our main paper, we apply the framework of \cite{arpit2017} to analyze 
learning dynamics of $S$ to reason why $S$ possesses  \textcolor{magenta}{\bf P1}, and therefore could be an effective proxy for $T$ under MI. 
We analyze generated samples $x_f$ from our T-ACGAN for 3 identities (IDs {\color{red} 71}, {\color{ForestGreen} 64}, {\color{Plum} 93}). 
We use 150 samples for each identity and show results for both the training set (top) and the validation set (bottom).
Note that $x_f$ analysis is relevant as generated samples are used in MI attacks.
{\bf (a)}: 
Different clusters and different distances from cluster centroids can be observed, suggesting 
patterns of face identities in some samples (easy samples) while
diverse
appearance in 
other samples (hard samples). 
We use distances from centroids to identify easy samples  
$x_f^e$ and hard samples $x_f^h$ (easy samples are indicated using a transparent blue circle for each ID in the visualization). 
Visualization of $x_f^e$ and $x_f^h$ in image space further demonstrates identity patterns in $x_f^e$
and diverse
appearance in $x_f^h$.
{\bf (b)}:
Similar to \cite{arpit2017}, we observe that 
$x_f^e$ and $x_f^h$ tend to have high and low likelihood under $T$ ($P_T$) resp. This is shown using 500k training data (top) and 500k validation data (bottom).
{\bf (c)}:  
We track likelihood under $S$ ($P_S$) for 
$x_f^e$ and $x_f^h$ during the training of $S$.
As training progresses, $P_S$ of $x_f^e$ and $x_f^h$ improve in general, and samples move up vertically (note that $P_T$ of samples do not change).
Consistent with the 
{\em 
``DNNs Learn Patterns First''}
finding
in \cite{arpit2017}, $S$ learns general identity patterns first to fit the easy samples.
Therefore, $P_S$ of $x_f^e$ improve at a faster pace in the training, and many of them achieve high $P_S$
at epoch = 200.
As $x_f^e$ tend to have high $P_T$, we observe property \textcolor{magenta}{\bf P1} in $S$.
% , achieves high likelihood on the easy samples (which also exhibit high likelihood under $T$). Hence we observe property P1 in $S$. 
For $x_f^h$ (many of them tend to have low $P_T$), 
%which exhibit low likelihood under $T$), 
it is uncommon for $S$ to achieve high likelihood on them as they do not fit easily to the pattern learned by $S$. 
Best viewed in color.
}
\label{fig_supp:training-dynamics-setup2}
\vspace{-0.5cm}
\end{figure}

\begin{figure}[!h]
% \vspace{-0.5cm}
\begin{adjustbox}{width=1.0\textwidth,center}
\begin{tabular}{c}
    \includegraphics[width=0.99\textwidth]{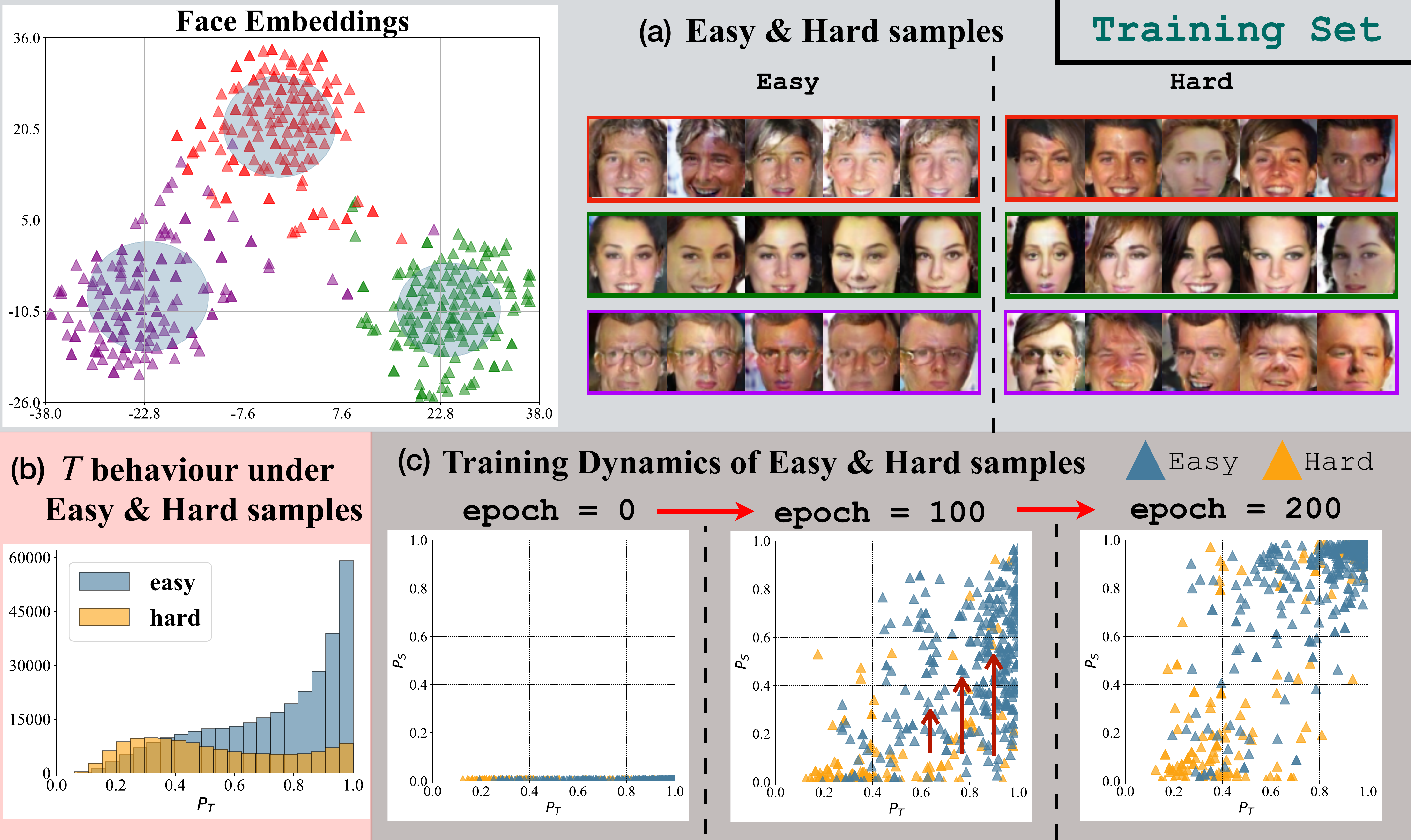} \\
    \hline
    \includegraphics[width=0.99\textwidth]{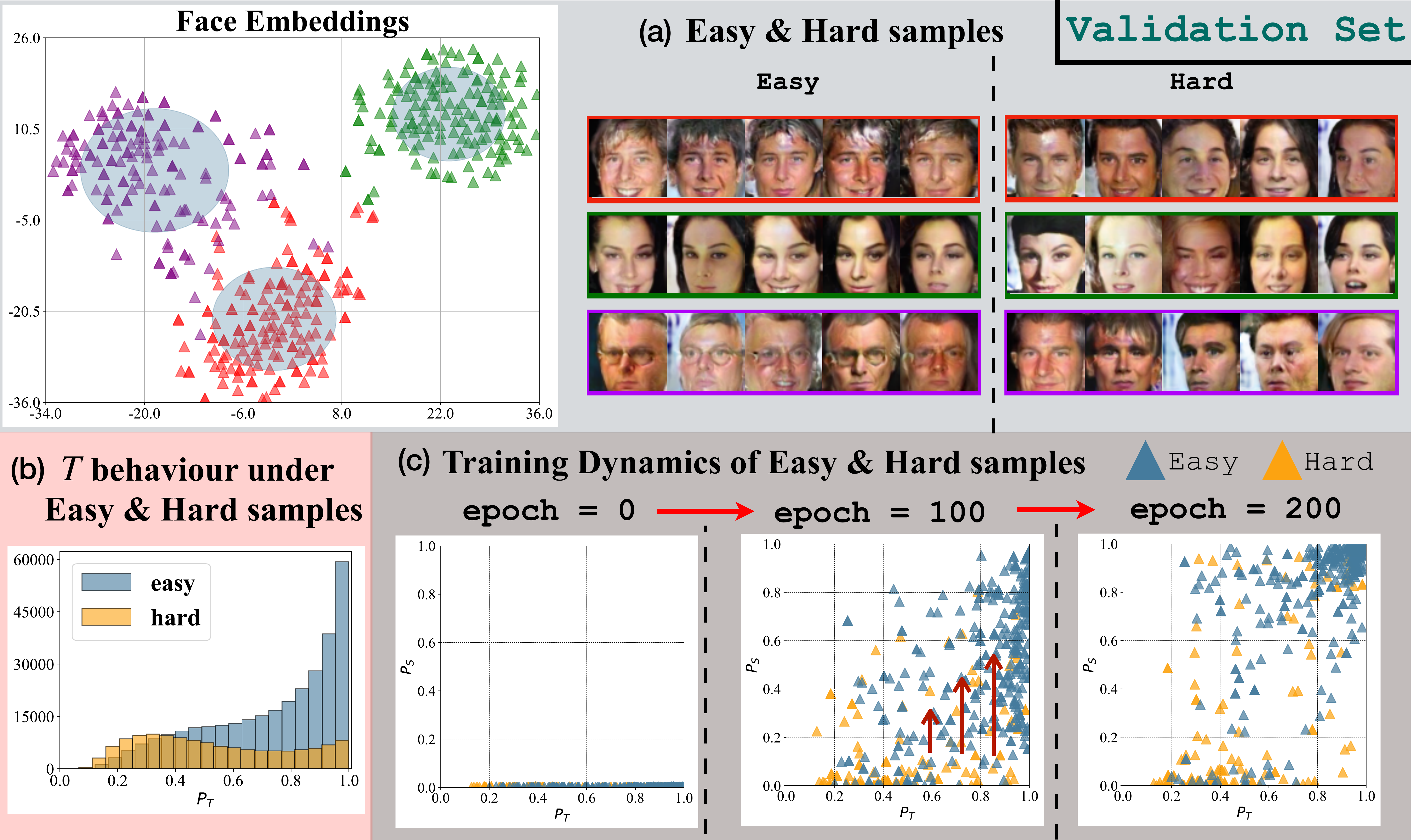}
\end{tabular}
\end{adjustbox}
\vspace{-0.35cm}
\caption{
\small
We use $\D_{priv}$ = CelebA \cite{liu2015deep}, $\D_{pub}$ = CelebA \cite{liu2015deep}, $T$ = FaceNet64, $S$ = DenseNet-161. 
The face embeddings are extracted using publicly available SOTA face recognition models \href{https://github.com/ageitgey/face_recognition}{here}. 
Similar to our main paper, we apply the framework of \cite{arpit2017} to analyze 
learning dynamics of $S$ to reason why $S$ possesses  \textcolor{magenta}{\bf P1}, and therefore could be an effective proxy for $T$ under MI. 
We analyze generated samples $x_f$ from our T-ACGAN for 3 identities (IDs {\color{red} 121}, {\color{ForestGreen} 95}, {\color{Plum} 163}). 
We use 150 samples for each identity and show results for both the training set (top) and the validation set (bottom).
Note that $x_f$ analysis is relevant as generated samples are used in MI attacks.
{\bf (a)}: 
Different clusters and different distances from cluster centroids can be observed, suggesting 
patterns of face identities in some samples (easy samples) while
diverse
appearance in 
other samples (hard samples). 
We use distances from centroids to identify easy samples  
$x_f^e$ and hard samples $x_f^h$ (easy samples are indicated using a transparent blue circle for each ID in the visualization). 
Visualization of $x_f^e$ and $x_f^h$ in image space further demonstrates identity patterns in $x_f^e$
and diverse
appearance in $x_f^h$.
{\bf (b)}:
Similar to \cite{arpit2017}, we observe that 
$x_f^e$ and $x_f^h$ tend to have high and low likelihood under $T$ ($P_T$) resp. This is shown using 500k training data (top) and 500k validation data (bottom).
{\bf (c)}:  
We track likelihood under $S$ ($P_S$) for 
$x_f^e$ and $x_f^h$ during the training of $S$.
As training progresses, $P_S$ of $x_f^e$ and $x_f^h$ improve in general, and samples move up vertically (note that $P_T$ of samples do not change).
Consistent with the 
{\em 
``DNNs Learn Patterns First''}
finding
in \cite{arpit2017}, $S$ learns general identity patterns first to fit the easy samples.
Therefore, $P_S$ of $x_f^e$ improve at a faster pace in the training, and many of them achieve high $P_S$
at epoch = 200.
As $x_f^e$ tend to have high $P_T$, we observe property \textcolor{magenta}{\bf P1} in $S$.
% , achieves high likelihood on the easy samples (which also exhibit high likelihood under $T$). Hence we observe property P1 in $S$. 
For $x_f^h$ (many of them tend to have low $P_T$), 
%which exhibit low likelihood under $T$), 
it is uncommon for $S$ to achieve high likelihood on them as they do not fit easily to the pattern learned by $S$. 
Best viewed in color.
}
\label{fig_supp:training-dynamics-setup3}
\vspace{-0.5cm}
\end{figure}

\begin{figure}[t!]
\vspace{-0.1cm}
\begin{adjustbox}{width=1.0\textwidth,center}
\begin{tabular}{c}
    \includegraphics[width=0.99\textwidth]{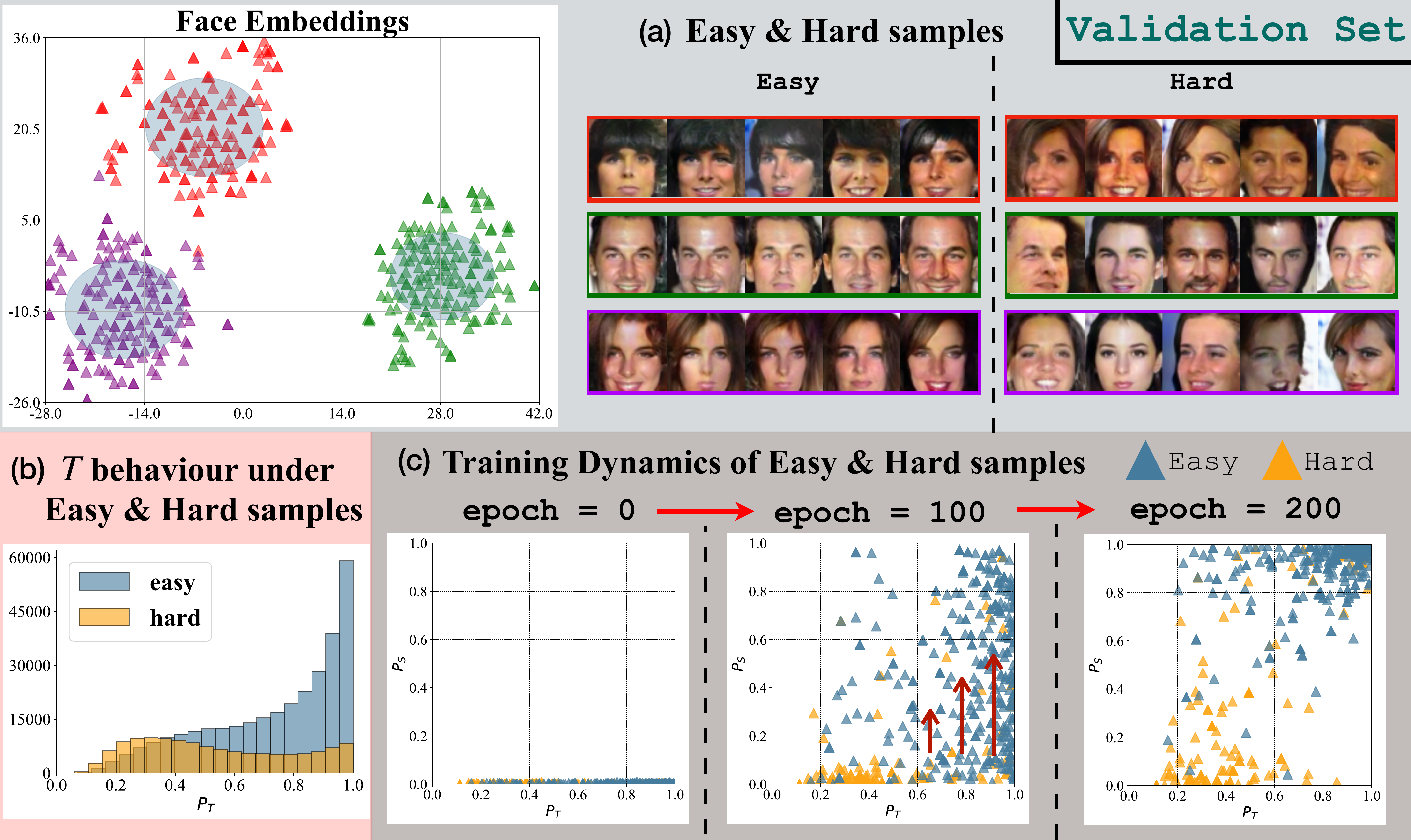}
\end{tabular}
\end{adjustbox}
\vspace{-0.35cm}
\caption{
\small
We use $\D_{priv}$ = CelebA \cite{liu2015deep}, $\D_{pub}$ = CelebA \cite{liu2015deep}, $T$ = FaceNet64, $S$ = DenseNet-161. 
The face embeddings are extracted using publicly available SOTA face recognition models \href{https://github.com/ageitgey/face_recognition}{here}. 
Similar to our main paper, we apply the framework of \cite{arpit2017} to analyze 
learning dynamics of $S$ to reason why $S$ possesses  \textcolor{magenta}{\bf P1}, and therefore could be an effective proxy for $T$ under MI. 
We analyze generated samples $x_f$ from our T-ACGAN for 3 identities (IDs {\color{red} 20}, {\color{ForestGreen} 16}, {\color{Plum} 36}). 
We use 150 samples for each identity and show results for the validation set. The training set results are already shown in the {\color{red} main paper Fig. 2}.
Note that $x_f$ analysis is relevant as generated samples are used in MI attacks.
{\bf (a)}: 
Different clusters and different distances from cluster centroids can be observed, suggesting 
patterns of face identities in some samples (easy samples) while
diverse
appearance in 
other samples (hard samples). 
We use distances from centroids to identify easy samples  
$x_f^e$ and hard samples $x_f^h$ (easy samples are indicated using a transparent blue circle for each ID in the visualization). 
Visualization of $x_f^e$ and $x_f^h$ in image space further demonstrates identity patterns in $x_f^e$
and diverse
appearance in $x_f^h$.
{\bf (b)}:
Similar to \cite{arpit2017}, we observe that 
$x_f^e$ and $x_f^h$ tend to have high and low likelihood under $T$ ($P_T$) resp. This is shown using 500k training data (top) and 500k validation data.
{\bf (c)}:  
We track likelihood under $S$ ($P_S$) for 
$x_f^e$ and $x_f^h$ during the training of $S$.
As training progresses, $P_S$ of $x_f^e$ and $x_f^h$ improve in general, and samples move up vertically (note that $P_T$ of samples do not change).
Consistent with the 
{\em 
``DNNs Learn Patterns First''}
finding
in \cite{arpit2017}, $S$ learns general identity patterns first to fit the easy samples.
Therefore, $P_S$ of $x_f^e$ improve at a faster pace in the training, and many of them achieve high $P_S$
at epoch = 200.
As $x_f^e$ tend to have high $P_T$, we observe property \textcolor{magenta}{\bf P1} in $S$.
% , achieves high likelihood on the easy samples (which also exhibit high likelihood under $T$). Hence we observe property P1 in $S$. 
For $x_f^h$ (many of them tend to have low $P_T$), 
%which exhibit low likelihood under $T$), 
it is uncommon for $S$ to achieve high likelihood on them as they do not fit easily to the pattern learned by $S$. 
Best viewed in color.
}
\label{fig_supp:training_dynamics_setup0}
\vspace{-0.5cm}
\end{figure}

% Part 2
\textbf{Why would $S$ possess  \textcolor{magenta}{P1}?} 
We provide additional empirical results using training and validation sets to support why $S$ possesses  \textcolor{magenta}{\bf P1} using the framework by \cite{arpit2017}.
% {\color{blue} Keshik: Will add the experiment details soon. I will also add the CelebA/ FFHQ 1(e) results.}
We use publicly available SOTA face recognition model(s)
\footnote{\url{https://github.com/ageitgey/face_recognition}} to extract face embeddings (128-dimensional) for analysis.
We use the following setup for analysis: $\D_{priv}$ = CelebA \cite{liu2015deep}, $\D_{pub}$ = CelebA \cite{liu2015deep}, $T$ = FaceNet64, $S$ = DenseNet-161.
Based on the distance from the face-embedding centroid for each identity, we consider the closest 70\% of samples as easy samples, and the remaining 30\% samples as hard samples 
\footnote{Note that the 70:30 selection of easy:hard samples has no effect to our algorithm; in fact our algorithm does not need explicit separation of easy/hard samples. 
Here in this discussion,  we separate easy and hard samples only to ease our illustration of 
{\em different pace of $P_S$ improvement among the samples},  which results in most samples with $P_S$ > 0.9 having high $P_T$.}.
The training dynamic results for easy and hard samples for 3 sets of randomly chosen identities are shown in Fig. \ref{fig_supp:training-dynamics-setup1}, \ref{fig_supp:training-dynamics-setup2} and \ref{fig_supp:training-dynamics-setup3}, for both training and validation sets.
We also show the training dynamics for the validation set corresponding to the main paper analysis results in Fig. \ref{fig_supp:training_dynamics_setup0}.

%% ==================Analysis Visualization (Training Setup 1) =============

%% ==================Analysis Visualization (Training Setup 2) =============

%% ==================Analysis Visualization (Training Setup 3) =============

%% ==================Analysis Visualization (Validation Setup 0) =============

% Decision knowledge transfer Analysis
\subsection{Decision knowledge transfer to T-ACGAN during training}
% \begin{wrapfigure}[50]{l}{0.55\textwidth}
\begin{figure}[t!]
\centering
% \vspace{-0.5cm}
    \includegraphics[width=0.5\textwidth]{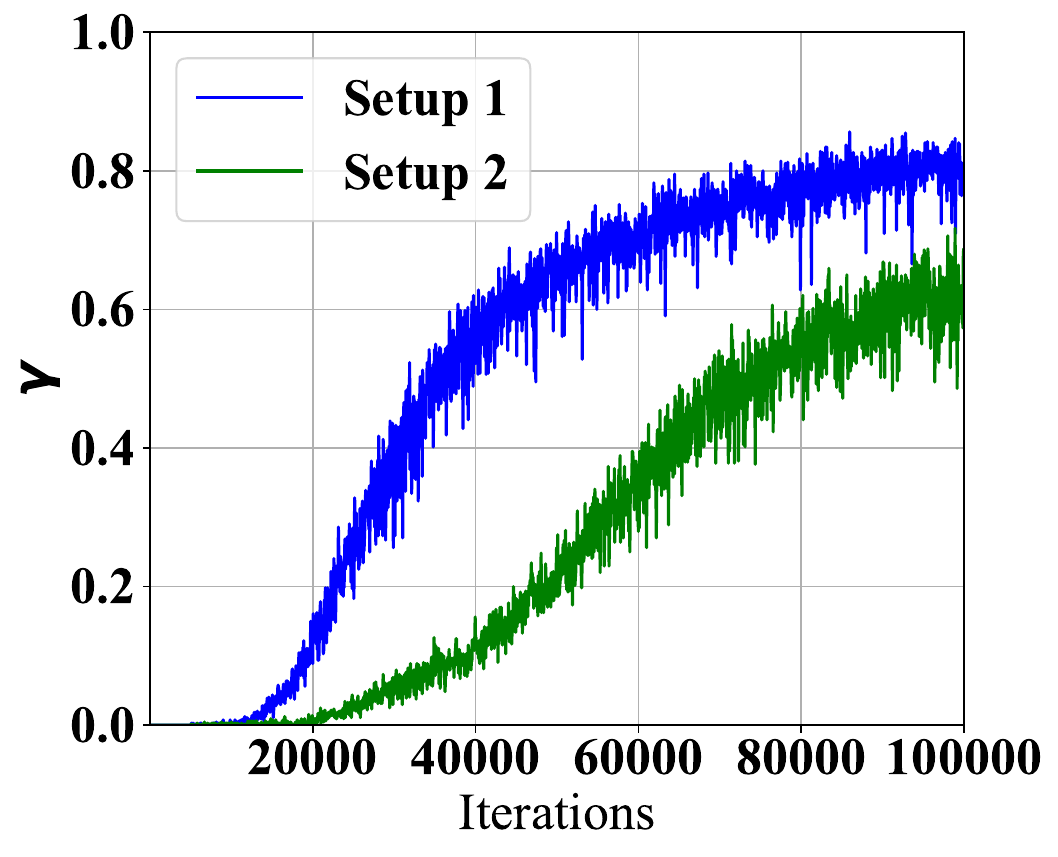}
% \vspace{-0.2cm}
\caption{
\small
We report $\gamma$ tracking results during T-ACGAN training for 2 setups. 
Our T-ACGAN is trained for 100k iterations. 
In both setups, $\bullet$ $\gamma$ starts low ($\gamma \approx 0$ for iterations less than 5000)
$\bullet$ With increasing iterations, $\gamma$ increases indicating adequate decision knowledge transfer from the target model $T$ to T-ACGAN. 
We remark that {\bf \color{ForestGreen} Setup 2} has lower $\gamma$ in general compared to {\bf \color{Blue} Setup 1} due to a large distribution shift between public data and private data.
}
\label{fig_supp:gamma}
%\end{figure}
% \end{wrapfigure}

\end{figure}In this section, we provide additional analysis to support that the target model, $T$'s, decision knowledge is adequately transferred to our T-ACGAN during training. 
Following the definition in Sec. 4.3 (main paper), $x_{f}=G(z, y)$, $\tilde{y}=T(x_f)$, 
let $\gamma$ be the percentage of samples with $y$ the same as $\tilde{y}$ in a batch of samples. 
%let the agreement between $y$ and $\tilde{y}$ be $\gamma=\frac{I[y==\tilde{y}]}{||y||}$.
In particular, we track $\gamma$ throughout our T-ACGAN training. 
Initially, we expect $\gamma$ to be low and with increasing training iterations, we expect $\gamma$ to increase indicating adequate decision knowledge transfer from the target model $T$.
We report $\gamma$ tracking results for 2 experiment setups in Fig. \ref{fig_supp:gamma}.
$\bullet$ {\bf \color{Blue} Setup 1:} We use $\D_{priv}$ = CelebA \cite{liu2015deep}, $\D_{pub}$ = CelebA \cite{liu2015deep}, $T$ = FaceNet64. 
$\bullet$ {\bf \color{ForestGreen} Setup 2:} We use $\D_{priv}$ = CelebA \cite{liu2015deep}, $\D_{pub}$ = FFHQ \cite{karras2019style}, $T$ = FaceNet64. 
We remark that batch size=128
% $||y||=128$ 
as we track $\gamma$ for every batch.
% i.e.: batch size=128.
We train T-ACGAN for 100k iterations.
As one can observe, $\gamma$ starts low and gradually increases during T-ACGAN training indicating adequate knowledge transfer from $T$.
% {\color{blue} ============ Keshik: Updating ===============}
\clearpage

\setcounter{figure}{0} 
\setcounter{table}{0} 

\section{Additional results}

\subsection{Different white-box attacks with surrogate models}
In this section, we perform a set of experiments to demonstrate that the surrogate models trained using our proposed framework are versatile enough to be used with different white-box MI attacks.
%in Table \ref{tab:whitebox}. 
For this analysis, we use two SOTA white-box attacks, namely KEDMI \cite{chen2021knowledge} and PLGMI \cite{yuan2023pseudo}.
For each white-box attack, we train five different surrogate models using our proposed framework including:
%Specifically, we employ five surrogate models 
$C \circ D$, $S_{DN121}$ = Desenet-121, $S_{DN161}$ = Desenet-161, $S_{DN169}$ = Desenet-169, and $S_{en} =\{$Desenet-121, Desenet-161, Desenet-169$\}$, and then, evaluate the white-box MI attack performance on each of these surrogate models.
%, and evaluate their effectiveness using two white-box attacks including 
% GMI\cite{zhang2020secret}, KEDMI \cite{chen2021knowledge} and PLGMI \cite{yuan2023pseudo}.

% To apply GMI on our $S$, we utilize the general GAN provided by \cite{chen2021knowledge}.
In the case of KEDMI, we train a specific-GAN using our surrogate model $S$ using the official implementation \footnote{https://github.com/SCccc21/Knowledge-Enriched-DMI}. As for PLGMI \footnote{https://github.com/LetheSec/PLG-MI-Attack}, given that our T-ACGAN can serve as a replacement for the conditional GAN of PLGMI, we leverage our T-ACGAN to apply the PLGMI attack. 
It is noteworthy that the target classifier $T$ is not used during the attacks when we apply white-box attacks on our surrogate models.

We report the results in Table \ref{tab:whitebox}, utilizing the CelebA dataset setup. Our results demonstrate the effectiveness of our surrogate models using white-box MI attacks, and are consistent with the outcomes obtained using the target classifier $T$ in white-box attacks.

\begin{table}[ht]
\centering
\caption{We compare the attack results using different white-box attacks with five surrogate models. We use $T =$ FaceNet64, $\D_{priv} =$ CelebA, $\D_{pub} =$ CelebA.
The results show that our different designs of surrogate model perform well across different white-box attacks. Note that the white-box attack results on $T$ are included only as reference as our setup does not have access to $T$ parameters nor soft-label of $T$.
% , $C \circ D$, $S =$ Densenet-161,and $S_{en} = \{$Densenet-121, Densenet-161, Densenet-169$\}$.
% {\color{blue} Keshik: We do not use our T-ACGAN here for KEDMI, but we use T-ACGAN for PLGMI.}
} 
% \begin{adjustbox}{width=\textwidth,center}
\begin{tabular}{llcc}
\hline
\textbf{Attack} & \textbf{Model} & \textbf{Attack acc. $\uparrow$} & \textbf{KNN dt. $\downarrow$} \\ \hline
% \multirow{3}{*}{GMI} & $T$ \cite{zhang2020secret} & 26.20 $\pm$ 	4.66 &	1626.60 \\ \cmidrule(lr){2-4} 
% & $C \circ D$ & 34.67 $\pm$	6.58	&  1552.79 \\ 
% & $S$ & 32.27 $\pm$	6.42 & 1595.90 \\ 
% & $S_{en}$ & 44.27	$\pm$ 4.82	& 1501.92 \\ \hline
\multirow{6}{*}{KEDMI} & $T$ \cite{chen2021knowledge} & 81.13 $\pm$ 4.66  &	1298.63 \\ \cmidrule(lr){2-4} 
& $C \circ D$  & 42.07 $\pm$	3.46	& 1473.99 \\
 & $S_{DN121}$ & 62.93  $\pm$	4.67 &	1350.67  \\ 
 & $S_{DN161}$ & 65.07 $\pm$ 3.79 &	1351.07  \\ 
 & $S_{DN169}$ & 62.80	$\pm$ 4.45 &	1350.56 \\ 
& $S_{en}$ & 69.00	$\pm$ 4.03 & 1329.84 \\ \hline
\multirow{6}{*}{PLGMI} & $T$ \cite{yuan2023pseudo} & 99.00 $\pm$ 0.01 & 1103.03  \\  \cmidrule(lr){2-4} 
& $C \circ D$ & 81.00 $\pm$	4.79 & 1298.63 \\ 
 & $S_{DN121}$ & 92.27	$\pm$ 2.85 &	1208.55  \\ 
 & $S_{DN161}$ & 92.80 $\pm$ 2.59 & 1207.25 \\ 
 & $S_{DN169}$ & 92.33	$\pm$ 3.36 &	1206.15 \\ 
& $S_{en}$ & 93.93 $\pm$ 2.78 & 1181.72 \\ \hline
\end{tabular}
% \end{adjustbox}
\label{tab:whitebox}
\end{table}

\subsection{Different TACGAN architecture}

For a fair comparison with BREPMI \cite{kahla2022label}, we provide the experiment results by training a new T-ACGAN using the same architectures as the GAN used by BREPMI. For the discriminator (D), we apply max pooling and add a linear layer before the last layer for the classifier head. As for the generator (G), we retain the same architecture and replace batch normalization with conditional batch normalization. 

We report the results in Table \ref{tab:same_architecture}. Our results are better than BREPMI when using the same GAN architecture.

\begin{table}[t]
 \setlength{\tabcolsep}{2.5pt}
\caption{We conduct comprehensive 
comparison between our proposed method 
and existing SOTA BREPMI \cite{kahla2022label} using the same GAN architecture. Specifically, we evaluate the performance of our three proposed  designs 
of surrogate, namely $C \circ D$, $S$, and $S_{en}$, while BREPMI performs black-box search on $T$ directly.
We highlight the best results in each setup in \textbf{bold}.
}

\begin{adjustbox}{width=0.55\columnwidth,center}
\begin{tabular}{llllc}
\cmidrule(lr){1-5}
\multicolumn{1}{c}{\textbf{Setup}} & \multicolumn{2}{c}{\textbf{Attack}} & \multicolumn{1}{c}{\textbf{Attack acc. $\uparrow$}} & \multicolumn{1}{c}{\textbf{KNN dt. $\downarrow$}} \\ 
\cmidrule(lr){1-5}

\multirow{4}{2.8cm}{$T$ \hspace{0.48cm} = FaceNet64  $\D_{priv}$ = CelebA  $\D_{pub}$ \hspace{0.01cm} = CelebA}   & \multicolumn{2}{l}{BREPMI} & 73.93   $\pm $ 4.98 & 1284.41 \\ \cmidrule(lr){2-5}
 & \multirow{3}{*}{\textbf{LOKT}} & $C \circ D$ & 85.47	$\pm $ 2.95 &	1336.45 \\
 &  & $S$ & 90.73	$\pm $ 3.57 & 1251.16 \\
 &  & $S_{en}$ & \textbf{93.20 $\pm $	1.98}	& \textbf{1214.60} \\ 
\cmidrule(lr){1-5}

\multirow{4}{2.8cm}{$T$ \hspace{0.48cm} = IR152  $\D_{priv}$ = CelebA  $\D_{pub}$ \hspace{0.01cm} = CelebA}  & \multicolumn{2}{l}{BREPMI} & 71.47   $\pm $ 5.32 & 1277.23 \\ \cmidrule(lr){2-5}
 & \multirow{3}{*}{\textbf{LOKT}}  & $C   \circ D$ &  88.20 $\pm $ 	3.48	& 1304.05  \\
 & & $S$  &  92.27	$\pm $ 2.46	& 1236.87 \\
 & & $S_{en}$  &  \textbf{94.53	$\pm $ 2.34}	& \textbf{1214.38} \\ 
\cmidrule(lr){1-5}

\multirow{4}{2.8cm}{$T$ \hspace{0.48cm} = VGG16  $\D_{priv}$ = CelebA  $\D_{pub}$ \hspace{0.01cm} = CelebA}  & \multicolumn{2}{l}{BREPMI} & 57.40   $\pm $ 4.92 & 1376.94  \\ \cmidrule(lr){2-5}
 & \multirow{3}{*}{\textbf{LOKT}}  & $C   \circ D$   & 68.93	$\pm $ 4.23	& 1450.74 \\
 & & $S$  &  78.07	$\pm $ 2.91	& 1362.70 \\
 & & $S_{en}$  &  \textbf{82.80	$\pm $ 3.20}	& \textbf{1346.51} \\ 
\cmidrule(lr){1-5}

\multirow{4}{2.8cm}{$T$ \hspace{0.48cm} = FaceNet64  $\D_{priv}$ = CelebA  $\D_{pub}$ \hspace{0.01cm} = FFHQ}   & \multicolumn{2}{l}{BREPMI} & 43.00   $\pm $ 5.14 & 1470.55  \\ \cmidrule(lr){2-5}
 & \multirow{3}{*}{\textbf{LOKT}}  & $C   \circ D$ & 59.87	5.05	& 1509.09   \\
 & & $S$  &   67.20	$\pm $ 4.23	& 1467.62 \\
 & & $S_{en}$  &  \textbf{72.33	$\pm $ 3.30}	& \textbf{1454.43} \\ 
\cmidrule(lr){1-5}

\end{tabular}
\end{adjustbox}
\label{tab:same_architecture}
\vspace{-0.5cm}
\end{table}

\subsection{White-box attack results for reference}

We show the our proposed method and other SOTA white-box attacks including GMI \cite{zhang2020secret}, KEDMI \cite{chen2021knowledge}, PLGMI \cite{yuan2023pseudo}, and the SOTA label-only attack BREPMI \cite{kahla2022label} in Table \ref{tab:all_result} for reference.

% \textcolor{red}{\bf [Milad: What is the intention for this comparison? Is it an ablation? Our approach can be considered as PLGMI attack with an estimated target model. Then, why we are comparing with GMI and KEDMI when we do not have attack performance on S?]}

% , Table \ref{tab:pubfig_facescrub}, and Table \ref{tab:ffhq}.

\begin{table}[ht]
%\vspace{-0.5cm}
\setlength{\tabcolsep}{1.5pt}
\centering
\caption{
We evaluate the performance of our label-only attack method across various experimental setups. For reference, we also include our results against three state-of-the-art (SOTA) white-box attacks, namely GMI \cite{zhang2020secret}, KEDMI \cite{chen2021knowledge}, PLGMI \cite{yuan2023pseudo}, as well as the SOTA label-only attack BREPMI \cite{kahla2022label}.
The obtained results clearly demonstrate the effectiveness of our label-only attack method over BREPMI, while also achieving comparable performance with other white-box attacks. }
\begin{adjustbox}{width=1.0\columnwidth,center}
\begin{tabular}{ccccccccccc}
\hline
\multicolumn{5}{c}{\textbf{Label-only MI Attacks}} & \multicolumn{6}{c}{\textbf{White-box MI Attacks (for reference only)}} \\ \cmidrule(lr){1-5} \cmidrule(lr){6-11}
\multicolumn{3}{c}{\textbf{LOKT}} & \multicolumn{2}{c}{\textbf{BREPMI \cite{kahla2022label}}} & \multicolumn{2}{c}{\textbf{GMI \cite{zhang2020secret}}} & \multicolumn{2}{c}{\textbf{KEDMI \cite{chen2021knowledge}}} & \multicolumn{2}{c}{\textbf{PLGMI \cite{yuan2023pseudo}}} \\ \cmidrule(lr){1-3} \cmidrule(lr){4-5} \cmidrule(lr){6-7} \cmidrule(lr){8-9} \cmidrule(lr){10-11} 
\multicolumn{1}{c}{S} 
& \multicolumn{1}{c}{\textbf{Attack acc. $\uparrow$}}
& \multicolumn{1}{c}{\textbf{KNN dt. $\downarrow$}} 
& \multicolumn{1}{c}{\textbf{Attack acc. $\uparrow$}}
& \multicolumn{1}{c}{\textbf{KNN dt. $\downarrow$}} 
& \multicolumn{1}{c}{\textbf{Attack acc. $\uparrow$}}
& \multicolumn{1}{c}{\textbf{KNN dt. $\downarrow$}}
& \multicolumn{1}{c}{\textbf{Attack acc. $\uparrow$}}
& \multicolumn{1}{c}{\textbf{KNN dt. $\downarrow$}} 
& \multicolumn{1}{c}{\textbf{Attack acc. $\uparrow$}}
& \multicolumn{1}{c}{\textbf{KNN dt. $\downarrow$}} \\ \hline

\multicolumn{11}{c}{ \textbf{$T =$ FaceNet64, $\D_{priv} =$ CelebA, $\D_{pub} =$ CelebA}} \\ \hline
\textbf{$C \circ D$} & 81.00 $\pm$ 4.79 & 1298.63
& \multirow{3}{*}{73.93 $\pm$ 4.98 } & \multirow{3}{*}{1284.41} 
& \multirow{3}{*}{26.20 $\pm$ 4.66 } & \multirow{3}{*}{1626.60} 
& \multirow{3}{*}{81.13 $\pm$ 4.66 } & \multirow{3}{*}{1247.91} 
& \multirow{3}{*}{99.00 $\pm$ 0.01} & \multirow{3}{*}{1103.03} \\ \cmidrule(lr){1-3}
\textbf{$S$} & 92.80 $\pm$ 2.59 & 1207.25 & & & & & & & & \\ \cmidrule(lr){1-3}
\textbf{$S_{en}$} & 93.93 $\pm$ 2.78 & 1181.72 & & & & & & & & \\ \hline

\multicolumn{11}{c}{ \textbf{$T =$ IR152, $\D_{priv} =$ CelebA, $\D_{pub} =$ CelebA}} \\ \hline
\textbf{$C \circ D$} & 72.07 $\pm$ 4.03 & 1358.94
& \multirow{3}{*}{71.47 $\pm$ 5.32 } & \multirow{3}{*}{1277.23} 
& \multirow{3}{*}{29.47 $\pm$ 4.70 } & \multirow{3}{*}{1609.57} 
& \multirow{3}{*}{79.87 $\pm$ 3.52 } & \multirow{3}{*}{1251.37} 
& \multirow{3}{*}{100.0 $\pm$ 0.00 } & \multirow{3}{*}{1026.71} \\ \cmidrule(lr){1-3}
\textbf{$S$} & 89.80 $\pm$ 2.33 & 1220.00 & & & & & & & & \\ \cmidrule(lr){1-3}
\textbf{$S_{en}$} & 92.13 $\pm$ 2.06 & 1206.78 & & & & & & & & \\ \hline

\multicolumn{11}{c}{ \textbf{$T =$ VGG16, $\D_{priv} =$ CelebA, $\D_{pub} =$ CelebA}} \\ \hline
\textbf{$C \circ D$} & 71.33 $\pm$ 4.39 & 1364.47
& \multirow{3}{*}{57.40 $\pm$ 4.92 } & \multirow{3}{*}{1376.94} 
& \multirow{3}{*}{18.07 $\pm$ 4.44 } & \multirow{3}{*}{1705.04} 
& \multirow{3}{*}{74.07 $\pm$ 4.21 } & \multirow{3}{*}{1290.81} 
& \multirow{3}{*}{97.00 $\pm$ 0.01} & \multirow{3}{*}{1120.61} \\ \cmidrule(lr){1-3}
\textbf{$S$} & 85.60 $\pm$ 3.03 & 1252.09 & & & & & & & & \\ \cmidrule(lr){1-3}
\textbf{$S_{en}$} & 87.27 $\pm$ 1.97 & 1246.71 & & & & & & & & \\ \hline

\multicolumn{11}{c}{ \textbf{$T =$ BiDO-HSIC \cite{peng2022bilateral}, $\D_{priv} =$ CelebA, $\D_{pub} =$ CelebA}} \\ \hline
\textbf{$C \circ D$} & 45.73 $\pm$ 5.94 & 1493.48 
& \multirow{3}{*}{37.40 $\pm$ 3.66 } & \multirow{3}{*}{1500.45} 
& \multirow{3}{*}{5.93 $\pm$ 1.85 } & \multirow{3}{*}{1930.52} 
& \multirow{3}{*}{42.80 $\pm$ 4.58 } & \multirow{3}{*}{1478.32} 
& \multirow{3}{*}{87.53 $\pm$ 3.08 } & \multirow{3}{*}{1237.41} \\ \cmidrule(lr){1-3}
\textbf{$S$} & 58.53 $\pm$ 4.87 & 1427.22 & & & & & & & & \\ \cmidrule(lr){1-3}
\textbf{$S_{en}$} & 60.73 $\pm$ 3.07 & 1395.93 & & & & & & & & \\ \hline

\multicolumn{11}{c}{ \textbf{$T =$ FaceNet64, $\D_{priv} =$ Facescrub, $\D_{pub} =$ Facescrub }} \\ \hline
\textbf{$C \circ D$} & 45.70	$\pm$ 4.00 & 1296.29
& \multirow{3}{*}{40.20 $\pm$ 6.60 } & \multirow{3}{*}{1236.40} 
& \multirow{3}{*}{14.60 $\pm$ 3.70 } & \multirow{3}{*}{1599.67} 
& \multirow{3}{*}{55.20 $\pm$ 4.61 } & \multirow{3}{*}{1193.41} 
& \multirow{3}{*}{92.50 $\pm$ 2.91 } & \multirow{3}{*}{1012.74} \\ \cmidrule(lr){1-3}
\textbf{$S$} & 53.20 $\pm$ 5.29 & 1280.70 & & & & & & & & \\ \cmidrule(lr){1-3}
\textbf{$S_{en}$} & 58.60 $\pm$ 4.86 & 1225.13 & & & & & & & & \\ \hline

\multicolumn{11}{c}{ \textbf{$T =$ FaceNet64, $\D_{priv} =$ Pubfig83, $\D_{pub} =$ Pubfig83}} \\ \hline
\textbf{$C \circ D$} & 74.80 $\pm$ 5.93 &	924.58 
& \multirow{3}{*}{55.60 $\pm$ 4.34 } & \multirow{3}{*}{1012.83}
& \multirow{3}{*}{16.40 $\pm$ 4.77 } & \multirow{3}{*}{1338.61} 
& \multirow{3}{*}{66.00 $\pm$ 4.00 } & \multirow{3}{*}{1031.86} 
& \multirow{3}{*}{99.60 $\pm$ 0.89 } & \multirow{3}{*}{832.07} \\ \cmidrule(lr){1-3}
\textbf{$S$} & 61.60 $\pm$ 3.58	
& 995.08 & & & & & & & & \\ \cmidrule(lr){1-3}
\textbf{$S_{en}$} & 80.00	$\pm$ 3.16
 & 883.52 & & & & & & & & \\ \hline

\multicolumn{11}{c}{ \textbf{$T =$ FaceNet64, $\D_{priv} =$ CelebA, $\D_{pub} =$ FFHQ}} \\ \hline
\textbf{$C \circ D$} & 43.27 $\pm$ 3.53 & 1516.18
& \multirow{3}{*}{43.00 $\pm$ 5.14 } & \multirow{3}{*}{1470.55} 
& \multirow{3}{*}{11.00 $\pm$ 4.64 } & \multirow{3}{*}{1750.74} 
& \multirow{3}{*}{54.20 $\pm$ 5.16 } & \multirow{3}{*}{1443.44} 
& \multirow{3}{*}{95.00 $\pm$ 0.04} & \multirow{3}{*}{1241.41} \\ \cmidrule(lr){1-3}
\textbf{$S$} & 59.13 $\pm$ 2.77 & 1437.86 & & & & & & & & \\ \cmidrule(lr){1-3}
\textbf{$S_{en}$} & 62.07 $\pm$ 3.89 & 1428.04 & & & & & & & & \\ \hline

\multicolumn{11}{c}{\textbf{ $T =$ FaceNet64, $\D_{priv} =$ Facescrub, $\D_{pub} =$ FFHQ}} \\ \hline
\textbf{$C \circ D$} & 44.50	$\pm$ 5.98 & 1403.73
& \multirow{3}{*}{37.30 $\pm$ 3.99} & \multirow{3}{*}{ 1456.59} 
& \multirow{3}{*}{11.00 $\pm$ 3.63 } & \multirow{3}{*}{1864.71} 
& \multirow{3}{*}{50.80 $\pm$ 4.58} & \multirow{3}{*}{ 1337.96} 
& \multirow{3}{*}{89.10 $\pm$ 3.05 } & \multirow{3}{*}{1196.88} \\ \cmidrule(lr){1-3}
\textbf{$S$} & 47.20 $\pm$ 4.39 & 1404.85 & & & & & & & & \\ \cmidrule(lr){1-3}
\textbf{$S_{en}$} & 53.70 $\pm$ 4.57 & 1338.67 & & & & & & & & \\ \hline

\multicolumn{11}{c}{\textbf{ $T =$ FaceNet64, $\D_{priv} =$ Pubfig83, $\D_{pub} =$ FFHQ}} \\ \hline
\textbf{$C \circ D$} & 85.60	$\pm$ 2.61	 & 914.15
& \multirow{3}{*}{72.80 $\pm$ 3.90 } & \multirow{3}{*}{971.51} 
& \multirow{3}{*}{36.40 $\pm$ 5.55 } & \multirow{3}{*}{1199.00} 
& \multirow{3}{*}{84.00 $\pm$ 4.00 } & \multirow{3}{*}{891.21} 
& \multirow{3}{*}{100.0 $\pm$ 0.00 } & \multirow{3}{*}{787.57} \\ \cmidrule(lr){1-3}
\textbf{$S$} & 88.40 $\pm$ 2.97 & 920.99 & & & & & & & & \\ \cmidrule(lr){1-3}
\textbf{$S_{en}$} & 94.40 $\pm$ 3.85 & 862.24 & & & & & & & & \\ \hline

\end{tabular}
\end{adjustbox}
\label{tab:all_result}
\end{table}

\subsection{Model stealing}
One related area to training surrogate models for a target model is {\em model stealing} where an attacker aims to copy the performance of a target model.
In this section, we compare the performance of our proposed method for training surrogate models ---specifically designed for MI attacks--- with model stealing approaches.
More specifically, we apply the SOTA model stealing approach DFMS-HL \footnote{https://github.com/val-iisc/Hard-Label-Model-Stealing} \cite{sanyal2022towards} that only uses the hard labels to train the surrogate model $S$. 
%Specially, we use DFMS-HL \footnote{https://github.com/val-iisc/Hard-Label-Model-Stealing} \cite{sanyal2022towards} and 
We train two surrogate models $S = C \circ D$, and $S = $ Densenet-161 \cite{huang2017densely} using DFMS-HL and compare it with the trained surrogate models with the proposed approach.
Table \ref{tab:model_stealing} shows that the surrogate models trained with our proposed method can perform much better for MI attacks.

\begin{table}[ht]
\caption{The comparison on MI attacks results using our surrogate models and model stealing DFMS-HL \cite{sanyal2022towards}. Here we use PLGMI \cite{yuan2023pseudo}, $\D_{priv} =$ CelebA, $\D_{pub} =$ CelebA, $T =$ FaceNet64.}
\centering
\resizebox{0.75\linewidth}{!}{
\begin{tabular}{lcccc}
\hline
\multicolumn{1}{c}{\multirow{2}{*}{\textbf{S}}} & \multicolumn{2}{c}{\textbf{DFMS-HL \cite{sanyal2022towards}}} & \multicolumn{2}{c}{\textbf{LOKT}} \\ \cmidrule(lr){2-3} \cmidrule(lr){4-5}  
\multicolumn{1}{c}{} & \textbf{Attack Acc. $\uparrow$} & \textbf{KNN dt. $\downarrow$} & \textbf{Attack Acc. $\uparrow$} & \textbf{KNN dt. $\downarrow$} \\ \hline
$C \circ D$ & 14.00	 $\pm$	4.01 & 	1775.71 & 81.00	 $\pm$	4.79 & 	1298.63 \\
Densenet-161 &  67.13  $\pm$		3.67  & 	1411.45 & 92.80 $\pm$	2.59 & 	1207.25 \\ \hline
\end{tabular}
}
\label{tab:model_stealing}
\end{table}

% \clearpage
% % Keshik
\subsection{Architecture selection for Surrogate models}
Our proposed approach (casting label-only MI attack as white-box MI attack) allows the possibility for MI attackers to choose the surrogate model architecture(s).
In this section, we study the effect of model architectures on model accuracy and MI attack accuracy to empirically justify our use of DenseNet model variants \cite{huang2017densely} as surrogate models.
The details of this study is as follows:
We conduct MI attacks on three different model families including MobileNet (MobileNetV2 \cite{sandler2018mobilenetv2} and MobileNetV3-small/ large \cite{koonce2021mobilenetv3}), EfficientNet \cite{tan2019efficientnet} (EfficientNet-B0, EfficientNet-B1, EfficientNet-B2, EfficientNet-B3, EfficientNet-B4, EfficientNet-B7), and DenseNet \cite{huang2017densely} (DenseNet-121, DenseNet-161, DenseNet-169). 
The number of parameters for each model (in Millions) is given in Table \ref{table-supp:arch-number-of-parameters}. 
We first train these 12 model architectures using private dataset $\mathcal{D}_{priv}$ = CelebA \cite{liu2015deep} which contains 30,027 images/ 1,000 identities following the exact training protocol in \cite{chen2021knowledge}.

\begin{table}[!htp]
\centering
\caption{Number of parameters (in Millions) for different model architectures.}
\label{table-supp:arch-number-of-parameters}
\begin{adjustbox}{width=1.0\columnwidth,center}
\begin{tabular}{lccc|cccccc|ccc}\toprule
Model &Mobi\_small &Mobi\_large &Mobi\_v2 &Eff-B0 &Eff-B1 &Eff-B2 &Eff-B3 &Eff-B4 &Eff-B7 &Den-121 &Den-161 &Den-169 \\ \midrule
Parameters (M) &1.50 &3.93 &3.50 &5.29 &7.79 &9.11 &12.20 &19.30 &66.30 &11.10 &35.30 &19.10 \\
\bottomrule
\end{tabular}
\end{adjustbox}
\end{table}

\begin{figure}
\centering
% \begin{wrapfigure}[28]{l}{0.7\textwidth}
% \vspace{-0.5cm}
    \includegraphics[width=0.7\textwidth]{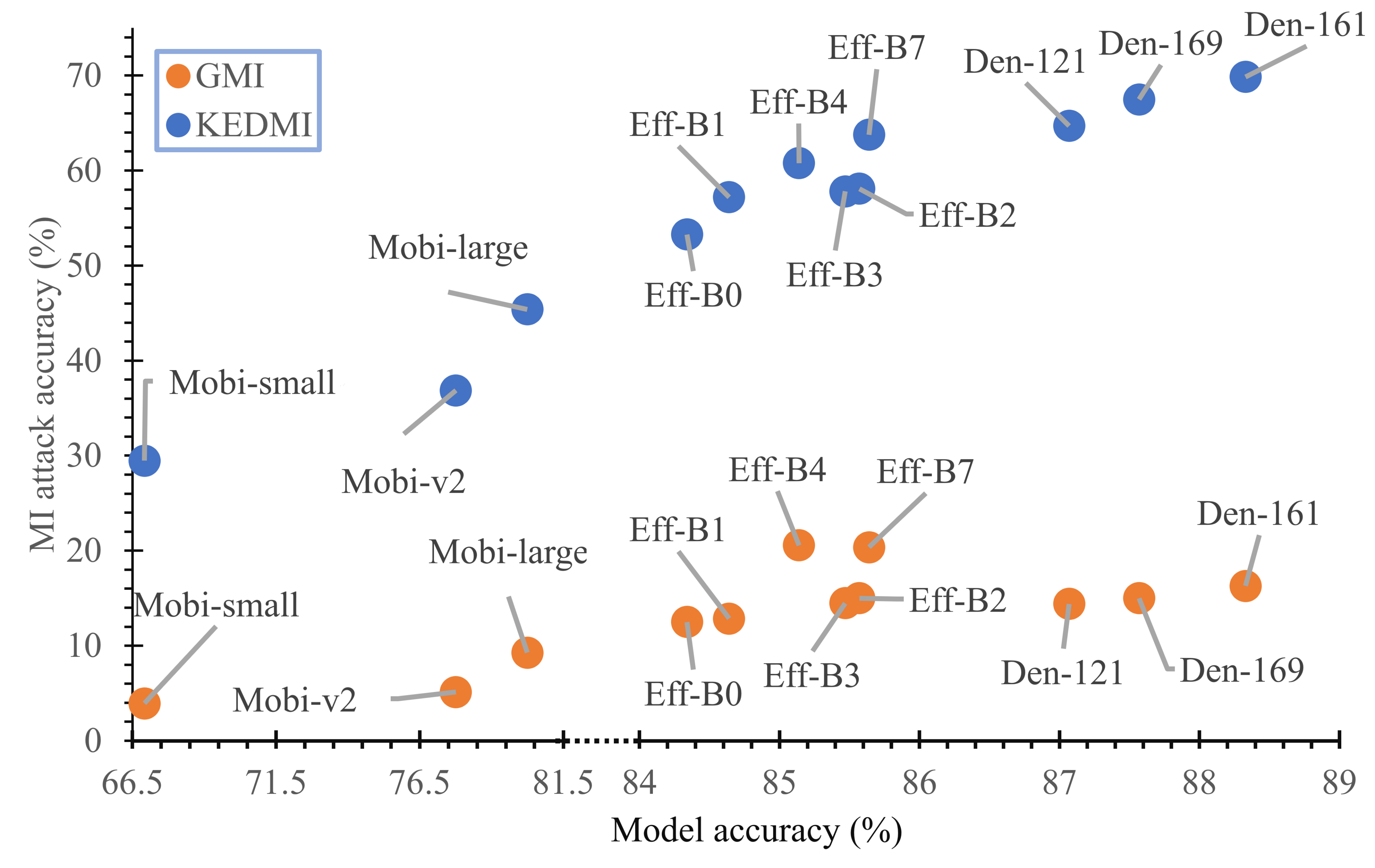}
% \vspace{-0.5cm}
\caption{
\small
\textit{Architecture selection for surrogate models:}
We report model accuracy and MI attack accuracy of 12 models from 3 model families namely, MobileNet (MobileNetV2 \cite{sandler2018mobilenetv2} and MobileNetV3-small/ large \cite{koonce2021mobilenetv3}), EfficientNet \cite{tan2019efficientnet} (EfficientNet-B0, EfficientNet-B1, EfficientNet-B2, EfficientNet-B3, EfficientNet-B4, EfficientNet-B7), and DenseNet \cite{huang2017densely} (DenseNet-121,DenseNet-161, DenseNet-169). 
The number of parameters for each model is included in Table \ref{table-supp:arch-number-of-parameters}. 
$\bullet$ We observe that compact models such as MobileNets obtain relatively lower model accuracy and lower MI Attack accuracy.
$\bullet$ We observe that larger models, i.e.: DenseNet models, achieve relatively higher model accuracy and higher MI Attack accuracy making them good candidates for surrogate models.
% We observe the low score in both model accuracy and MI attack accuracy in compact networks such as MobileNets. DensenNets have more parameters and achieve better MI attack accuracy.
}
\label{fig_supp:architectures}
\end{figure}
% \end{wrapfigure}

After training target models, we perform white-box MI attacks on these target models.
We use two popular white-box MI attacks namely GMI \cite{zhang2020secret} and KEDMI \cite{chen2021knowledge}. 
Following \cite{chen2021knowledge}, we use evaluation model $\mathnormal{E} = $ FaceNet \cite{cheng2017know}.
We report the model accuracy and MI attack accuracy in Fig. \ref{fig_supp:architectures}.
When comparing models within the same family, in general, we observe that architectures with more parameters achieve better model accuracy and are more susceptible to MI attacks (Higher MI Attack Acc).
Based on KEDMI \cite{chen2021knowledge} results obtained in this study, \textit{we use architectures from the DenseNet family\footnote{DenseNet-161 has more parameters than DenseNet-169  More details: \href{https://pytorch.org/vision/stable/models/densenet.html} as our surrogate model(s)}.}

% \textcolor{red}{\bf [Milad: Using a wrapped figure makes it harder to follow the text. As we do not have space limitation, having white space around a figure might be better than using wrapped figures.]}

% {\color{blue} ======== Keshik: Updating ===========}
% \clearpage
\section{Additional Reconstruction Results}
\begin{figure*}[ht]
\begin{adjustbox}{width=0.99\textwidth,center}
\begin{tabular}{c}
    \includegraphics[width=0.99\textwidth]{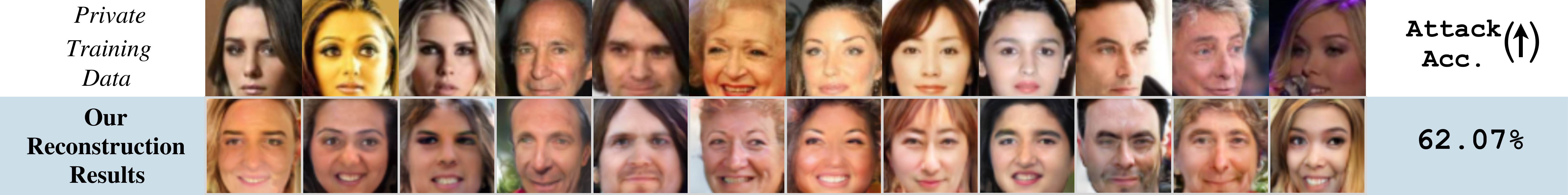}
\end{tabular}
\end{adjustbox}
% \vspace{-0.35cm}
\caption
{
We use $\mathcal{D}_{priv}$ = CelebA \cite{liu2015deep}, $\mathcal{D}_{pub}$ = FFHQ \cite{karras2019style}, $T$ = FaceNet64.
We show private data (top), \textit{our} reconstruction results (bottom) and Attack accuracy.
We remark that these results are obtained using $S_{en}$.
}
\label{fig-supp:celeba_ffhq_facenet64_samples}
% \vspace{-0.5cm}
\end{figure*}

\begin{figure*}[ht]
\begin{adjustbox}{width=0.99\textwidth,center}
\begin{tabular}{c}
    \includegraphics[width=0.99\textwidth]{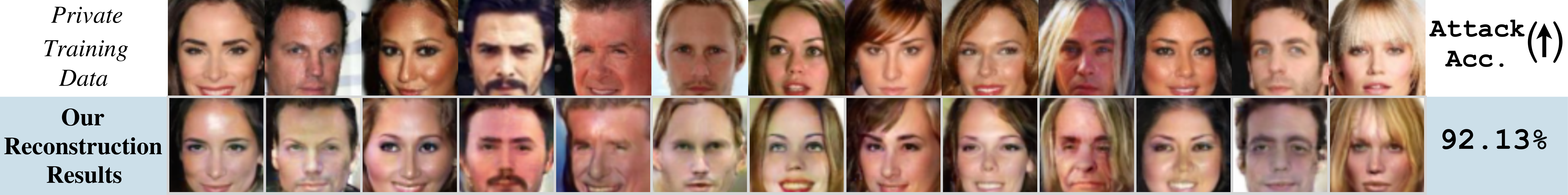}
\end{tabular}
\end{adjustbox}
% \vspace{-0.35cm}
\caption
{
We use $\mathcal{D}_{priv}$ = CelebA \cite{liu2015deep}, $\mathcal{D}_{pub}$ = CelebA \cite{liu2015deep}, $T$ = IR152 \cite{he2016deep}.
We show private data (top), \textit{our} reconstruction results (bottom) and Attack accuracy.
We remark that these results are obtained using $S_{en}$.
}
\label{fig-supp:celeba_celeba_ir152_samples}
% \vspace{-0.5cm}
\end{figure*}

\begin{figure*}[ht]
\begin{adjustbox}{width=0.99\textwidth,center}
\begin{tabular}{c}
    \includegraphics[width=0.99\textwidth]{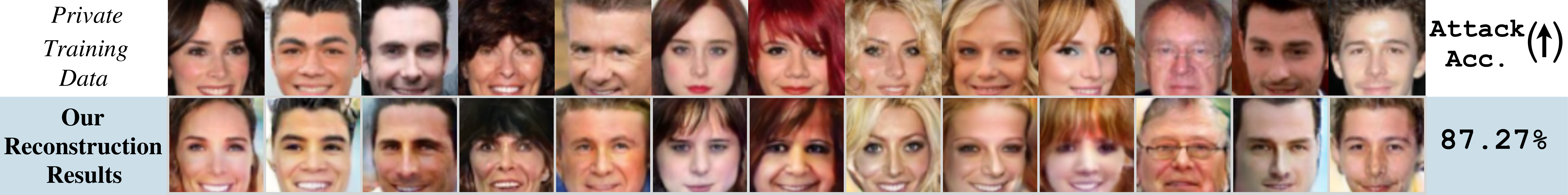}
\end{tabular}
\end{adjustbox}
% \vspace{-0.35cm}
\caption
{
We use $\mathcal{D}_{priv}$ = CelebA \cite{liu2015deep}, $\mathcal{D}_{pub}$ = CelebA \cite{liu2015deep}, $T$ = VGG16 \cite{simonyan2014very}.
We show private data (top), \textit{our} reconstruction results (bottom) and Attack accuracy.
We remark that these results are obtained using $S_{en}$.
}
\label{fig-supp:celeba_celeba_vgg16_samples}
% \vspace{-0.5cm}
\end{figure*}

In this section, we show reconstructed samples for 3 additional setups using our proposed method. 
We show cross-dataset MI results in Fig. \ref{fig-supp:celeba_ffhq_facenet64_samples} using FaceNet64 target model.
In addition, we also show results for 2 additional target models: IR152 \cite{he2016deep} and VGG16 \cite{simonyan2014very} in Fig. \ref{fig-supp:celeba_celeba_ir152_samples} and Fig. \ref{fig-supp:celeba_celeba_vgg16_samples} respectively. 

%%% ==================CelebA / FFHQ / GMI / FaceNet64 / Reconstruction=============

%%% ==================CelebA / CelebA / IR152 / Reconstruction=============

%%% ==================CelebA / CelebA / VGG16 / Reconstruction=============

\section{Experiment details/ Design choices}
% \subsection{Dataset}
We use three datasets including CelebA \cite{liu2015deep}, Facescrub \cite{ng2014data}, and Pubfig83 \cite{pinto2011scaling}. We further examinate the distribution shift by using FFHQ dataset \cite{karras2019style} which includes images that vary in terms of background, ethnicity, and age. 
Following \cite{chen2021knowledge,kahla2022label}, we divide CelebA into two datasets $\mathcal{D}_{priv}$ for training the target model $\mathcal{T}$ and $\mathcal{D}_{pub}$ for training GAN and surrogate models $\mathcal{C}$. The details of each dataset are summarized in Table \ref{tab:dataset}.

\begin{table}[ht]
\centering
\caption{Details of three datasets including CelebA \cite{liu2015deep}, Facescrub \cite{ng2014data}, and Pubfig83 \cite{pinto2011scaling}.}
% \begin{adjustbox}{width=0.95\columnwidth,center}
\begin{tabular}{lcccc}
\hline
\multirow{2}{*}{\textbf{Dataset}} & \multicolumn{2}{c}{\textbf{$\mathcal{D}_{priv}$}} & \multicolumn{2}{c}{\textbf{$\mathcal{D}_{pub}$}} \\ \cmidrule(lr){2-3} \cmidrule(lr){4-5}
 & \textbf{\# Target id} & \textbf{\# Images} & \textbf{\# Id} & \textbf{\# Images} \\ \hline
CelebA \cite{liu2015deep} & 1,000 & 30,027 & - & 30,000\\ \hline
Facescrub \cite{ng2014data} & 200 & 40,953 & 330 & 65,910 \\ \hline
Pubfig83 \cite{pinto2011scaling} & 50 & 8,145 & 33 & 5,693 \\ \hline
FFHQ \cite{karras2019style} & - & - & - & 70,000 \\ \hline
\end{tabular}
% \end{adjustbox}
\label{tab:dataset}
\end{table}

\section{Evaluation details}
Following \cite{kahla2022label}, we attack the first 300 out of 1000 labels in the experiments using CelebA dataset. In cases of Facescrub and Pubfig, we attack all the labels of the target classifier (200 and 50, respectively). As for the evaluation model, we use FaceNet which is trained on the private dataset and has higher resolution than the target classifier (image resolution 112x112). We remark that all pre-trained target/ evaluation models are released publicly by \cite{kahla2022label}, and we adopt these models in our experiments for fair comparison.

\subsection{T-ACGAN architecture}
We adopt the SNResnet architecture \cite{miyato2018cgans,miyato2018spectral} for our T-ACGAN. The architecture of the generator and the discriminator are as shown in Table \ref{tab:G} and \ref{tab:D}.
% \textcolor{red}{\bf [Milad: Shown in Table?]}

\begin{table}[ht]
\centering
\caption{Generator}
\begin{tabular}{lllll}
\hline
\textbf{Operation} & \textbf{Kernel} &\textbf{Strides} & \textbf{Feature maps} &\textbf{BN?} \\ \hline
Linear & N/A & N/A & 16384 &  \\ 
Convolution & 3x3 & 1x1 & 512 &  \\
Convolution & 3x3 & 1x1 & 512 & yes \\
Convolution & 1x1 & 1x1 & 512 &  \\
Convolution & 3x3 & 1x1 & 256 &  \\
Convolution & 3x3 & 1x1 & 256 & yes \\
Convolution & 1x1 & 1x1 & 256 &  \\
Convolution & 3x3 & 1x1 & 128 &  \\
Convolution & 3x3 & 1x1 & 128 & yes \\
Convolution & 1x1 & 1x1 & 128 &  \\
Convolution & 3x3 & 1x1 & 64 &  \\
Convolution & 3x3 & 1x1 & 64 & yes \\
Convolution & 1x1 & 1x1 & 64 & yes  \\
Convolution & 1x1 & 1x1 & 3 &  \\ \hline

\end{tabular}
\label{tab:G}
\end{table}

\begin{table}[ht]
\centering
\caption{Discriminator. $N$ is the number of classes.}
\begin{tabular}{llll}
\hline
\textbf{Operation} & \textbf{Kernel} & \textbf{Strides} & \textbf{Feature maps} \\ \hline
Convolution & 3x3 & 1x1 & 64 \\
Convolution & 3x3 & 1x1 & 64 \\
Convolution & 1x1 & 1x1 & 64 \\
Convolution & 3x3 & 1x1 & 64 \\
Convolution & 3x3 & 1x1 & 128 \\
Convolution & 1x1 & 1x1 & 128 \\
Convolution & 3x3 & 1x1 & 128 \\
Convolution & 3x3 & 1x1 & 256 \\
Convolution & 1x1 & 1x1 & 256 \\
Convolution & 3x3 & 1x1 & 256 \\
Convolution & 3x3 & 1x1 & 512 \\
Convolution & 1x1 & 1x1 & 512 \\
Convolution & 3x3 & 1x1 & 512 \\
Convolution & 3x3 & 1x1 & 1024 \\
Convolution & 1x1 & 1 & 1024 \\
Linear & N/A & N/A & 1 \\
Linear & N/A & N/A & N \\
\hline
\end{tabular}
\label{tab:D}
\end{table}

\subsection{Hyperparameters}
\textbf{Training T-ACGAN.}
% The number of iteration to train T-ACGAN are 100k, 25k, and 15k for CelebA \cite{liu2015deep}, Facescrub \cite{ng2014data}, and Pubfig83 \cite{pinto2011scaling}, respectively. We remark that is we train $G$ once and train $D$ five times.
The T-ACGAN model was trained using different numbers of iterations for CelebA \cite{liu2015deep}, Facescrub \cite{ng2014data}, and Pubfig83 \cite{pinto2011scaling} datasets. Specifically, we utilized 20k iterations for CelebA, 5k iterations for Facescrub, and 3k iterations for Pubfig83. It's important to note that during training, the generator $G$ was trained once while the discriminator $D$ was trained five times for each iteration. 
% Therefore, the reported numbers correspond to the iterations required to train the discriminator ($D$).
%For easier reference, we present equations (3) and (4) from the main paper. In these equations 
For T-ACGAN loss, including generator loss $\mathcal{L}_G$, and discriminator loss $\mathcal{L}_{D,C}$ (Eqn. (3) and (4) in the main paper), we select $\lambda_1 = 1.0$ and $\lambda_2 = 1.5$ for all experiments. This deliberate choice aims to enhance the learning process of both the generator and the discriminator by emphasizing the importance of conditional loss.

\begin{align*}
   \mathcal{L}_G = \lambda_1 E[\log P(s = Fake| x_{f})] - \lambda_2 E[\log P(c = {y} | x_f)]
 %\label{eq:acgan_G}  
\end{align*}

\begin{align*}
 %\begin{aligned}
 \mathcal{L}_{D,C} & =  - \lambda_1 [ E[\log P(s = Fake| x_{f})] - E[\log P(s = Real| x_p)] ] - \lambda_2 E[\log P(c = \ty | x_f)] 
%\end{aligned}
%\label{eq:tacgan_D}
\end{align*}

\textbf{Training surrogate models $S$ and $S_{en}$.}
As we mentioned in the main paper, to train additional surrogate models $S$ and $S_{en}$, we create a new synthetic dataset generated by our T-ACGAN.
%and label it by the target classifier $T$. 
Specially, we generate images using 500 pseudo labels for each class. These images are then labeled by the target classifier $T$.
To train $S$ and $S_{en}$, we use SGD optimizer with learning rate $lr = 0.1$, momentum 0.9 and weight decay $5 \times 10^{-4}$, and apply the CosineAnnealingLR scheduler \cite{LoshchilovH17}.

\textbf{Inversion.} To reconstruct the images, after training the surrogate model $S$, in the main experimental results, we apply PLGMI \cite{yuan2023pseudo} as the white-box MI attack on $S$ using our T-ACGAN. 
For this reconstruction, we use Adam optimizer with the learning rate $lr = 0.002$ and optimize in 600 iterations as \cite{yuan2023pseudo}.
For other experiments, when using KEDMI and GMI as white-box MI attacks on $S$, following \cite{chen2021knowledge}, we use SGD optimizer with learning rate $lr = 0.02$ and optimize in 2400 iterations.

\subsection{User study}

\begin{figure}[ht]
    \centering
    % \renewcommand\thefigure{A}
    % \begin{tabular}{c}
    
    \includegraphics[width=0.7\textwidth]{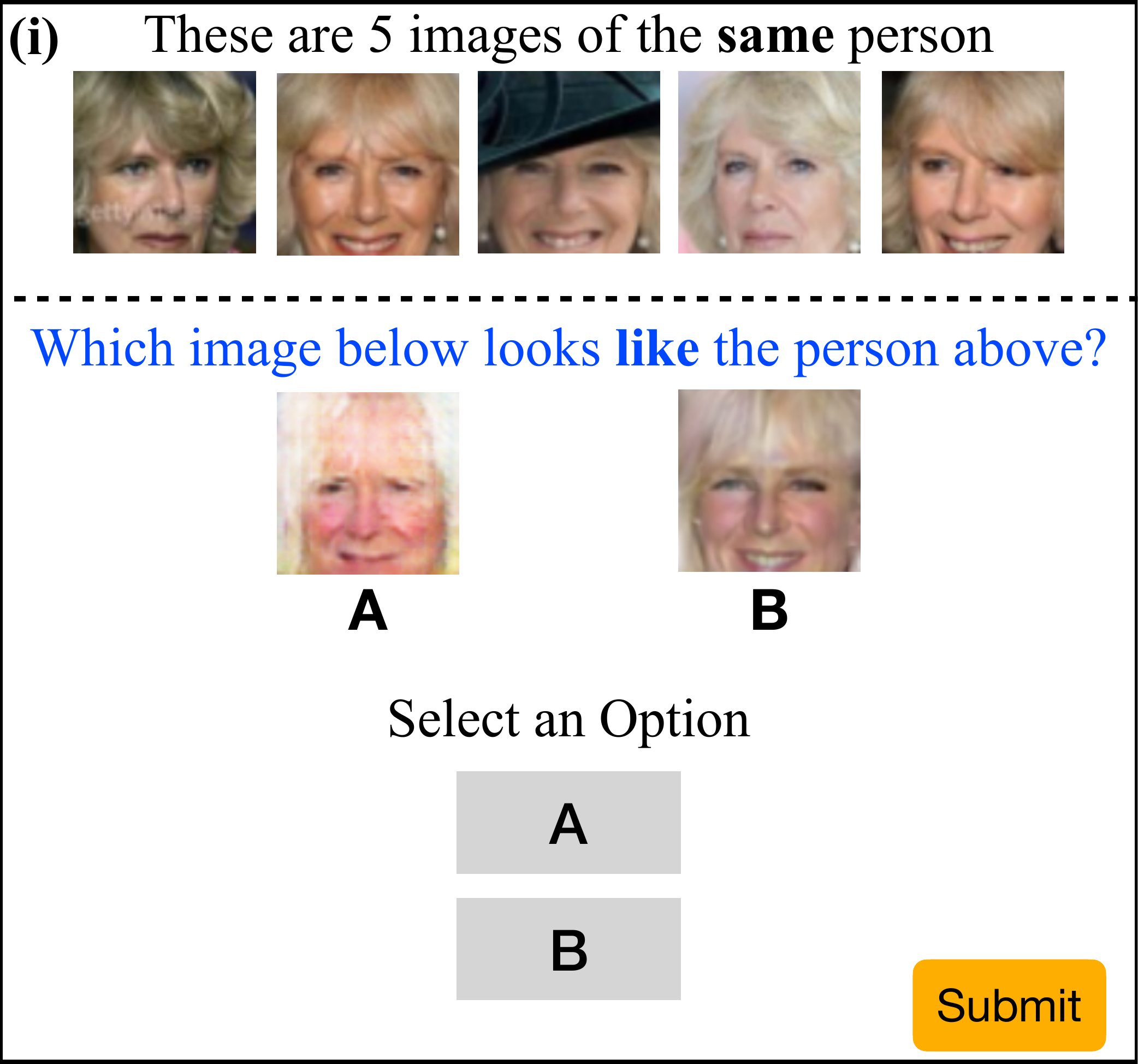} 
      
  \caption{
% {\em Human study unveiling serious privacy leakage under our proposed Label-only Model Inversion (MI) attack method}.
    \textbf{Human study setup/ user interface}: We follow the setup proposed by An \etal for human study.
  $\bullet$ In this setup, users are shown 5 real images of a person (identity) as reference.
  $\bullet$ Then users are required to
  compare the 5 real images with two inverted images: one from our method, the other from BREPMI.
  In the above example, A and B correspond to BREPMI and Ours respectively.
  The order is randomized for each task.
  Each user is given a maximum of 60 seconds per task, and each task is assigned to 10 unique users.
  Following \cite{an2022mirror}, we randomly select 50 identities, resulting in 1000 pairs.
  We use Amazon Mechanical Turk service (MTurk).  We use $D_{priv}$ = CelebA, $D_{pub}$ = CelebA, $T$ = FaceNet64. }
\label{fig:human_study}
\end{figure}

The user interface is shown in Figure \ref{fig:human_study}. The results are included in the main paper.

\section{Related works}
Model Inversion (MI) aims to extract/ reconstruct the private information about the training data through a trained model.
Depending on the level of information that can be accessed, MI attacks can be classified into three distinct categories: white-box attacks, black-box attacks, and label-only attacks.

\textbf{White-Box MI Attack.} In white-box attacks, the attacker is assumed to have complete access to the target model including model weights.
Therefore, the MI attack is usually formulated as optimizing an identification loss: 
\begin{equation}
\label{eq:white_box_spatial}
 x^* = \arg \min_x \mathcal{L}_{id}(x;y,T) 
\end{equation}
where $\mathcal{L}_{id}(x;y,T) = -\log \mathbb{P}_{T}(y|x)$, with $\mathbb{P}_{T}(y|x)$ denoting the probability (soft label) that the target model $T$ classifies input $x$ as label $y$.
When handling a high-dimensional input data $x$ like an image, performing the optimization (Eqn. \ref{eq:white_box_spatial}) in input space ends up with degraded results \cite{fredrikson2015model, zhang2020secret}.
To overcome this issue, recent white-box approaches \cite{zhang2020secret, chen2021knowledge, wang2021variational} constraint the search space into the manifold of  related public  images using a GAN.
More specifically, GMI \cite{zhang2020secret} proposes to train a GAN on a public dataset $\mathcal{D}_{pub}$, and perform the inversion step on the latent space of GAN:
\begin{equation}
\label{eq:white_box_latent}
 z^* = \arg \min_z \mathcal{L}_{id}(z;y,T) + \lambda \mathcal{L}_{prior}(z)
\end{equation}
where $z^*$ denotes the optimal latent code which is later used by GAN to generate the reconstructed sample, i.e., $x^* = G(z^*)$. In addition, $\mathcal{L}_{prior}=-D(G(z))$ measures the realness of the generated sample.
KEDMI \cite{chen2021knowledge} improves GMI by introducing inversion-specific GAN, and restoring a distribution of latent space instead of an optimal point. In addition, VMI \cite{wang2021variational} defines the variational inference in latent space. 
PLGMI \cite{yuan2023pseudo} uses the target classifier to produce pseudo label for public data and trains a conditional GAN (cGAN) to limit the search space.

\textbf{Black-Box MI Attack.}
In the black-box setup, the attackers have access to only model's output and confidence scores (soft labels) which is very limited compared to the white-box setup.
Due to this limitation, performing optimization discussed in Eqn. \ref{eq:white_box_spatial}, and \ref{eq:white_box_latent} become unfeasible in the black-box setup. Yang et al. \cite{yang2019neural} train an inversion model of the target model which serves as an encoder model specifically trained to produce the predicted score (soft labels). Simultaneously, the generator (decoder) is trained to generate the target image based on the predicted score of the inversion model.

\textbf{Label-Only MI Attack.}
Label-only MI attack relies solely on the final decision of the model, i.e., the predicted label, without any additional information about the model or the confidence score of the prediction. 
Kahla et. al \cite{kahla2022label} propose Boundary-Repelling Model Inversion (BREP-MI) to address the model inversion attack under label-only setup. Beginning by initializing a random point that is already classified into the target class, BREPMI  evaluates the model’s predicted labels based on other neighbor points in the latent space and estimate the direction to reach the target class’s centroid. 

In future work, we hope to explore different aspects of model inversion including multimodal learning, advanced knowledge transfer, data-centric applications and different types of generative models \cite{chandrasegaran2021closer, pmlr-v162-chandrasegaran22a, koh2023grounding, Chandrasegaran_2022_ECCV,khosla2020supervised, zhao2022few, abdollahzadeh2023survey}.

%  \textbf{White-box attacks.} 
%  In white-box attacks \cite{fredrikson2014privacy,fredrikson2015model,zhang2020secret,wang2021variational,chen2021knowledge,zhao2021exploiting,yuan2023pseudo}, the adversary possesses complete access to the target model, including its parameters and architecture. 
%  \textbf{Black-box attacks.}
% Black-box attacks involve a scenario where the adversary has limited access to the target model. In this setting, the attacker can only obtain the model's output, typically consisting of the predicted label and its associated confidence score, when provided with an input query. Black-box attacks present a more challenging task, requiring the adversary to devise strategies to extract private information based solely on the model's output.
%  \textbf{Label-only attacks.} In this attack, the adversary is only provided with the predicted hard label without any other information. Despite the limited information, Kahla et. al. \cite{kahla2022label} proposed BREPMI, a new approach to gradually moving away the decision boundary to find the reconstructed images.

 % (i) White-box attacks \cite{fredrikson2014privacy,fredrikson2015model,zhang2020secret,wang2021variational,chen2021knowledge,zhao2021exploiting,yuan2023pseudo} where the adversary is fully accessible to the target model, e.g, model parameters, model architectures, (ii) Black-box attacks where the adversary can receive the output (label and the confidence score), given an input query, (iii) Label-only attack \cite{kahla2022label} where adversary can get only the predicted hard label.
\setcounter{figure}{0} 
\setcounter{table}{0} 

\section{Additional information for checklist}

% \subsection{Amount of Compute}
\textbf{Amount of Compute.}
% \label{sec-supp:amount_compute}
The amount of compute in this project is reported in Table \ref{table-supp:compute}. 
We follow
NeurIPS guidelines to include the amount of compute for different experiments along with $CO_2$ emission.

\begin{table}[!h]
\caption{
Amount of compute in this project. The GPU hours include computations for initial
explorations / experiments to produce the reported values. CO2 emission values are
computed using \url{https://mlco2.github.io/impact/}
}
  \begin{adjustbox}{width=\textwidth}
  \begin{tabular}{lccc}\toprule
\textbf{Experiment} &\textbf{Hardware} &\textbf{GPU hours} &\textbf{Carbon emitted in kg} \\ \toprule
Main paper : Table 3 (Repeated 3 times) &RTX A5000 &306 &29.56 \\ \midrule
Main paper : Table 2 and Table 4 &RTX A5000 &50 &4.83 \\ \midrule
Main paper : Figure 1 / Figure 2 &RTX A5000 &4 &0.39 \\ \midrule
Supplementary : All additional analysis/ Ablation study &RTX A5000 &10 &0.97 \\ \midrule
% Supplementary : Extended Experiments &Tesla V100-PCIE (32 GB) &68 &6.12 \\ \midrule
% Supplementary : Ablation Study &Tesla V100-PCIE (32 GB) &14 &1.26 \\ \midrule
Additional Compute for Hyper-parameter tuning &RTX A5000 &24 &2.32 \\ \midrule
\textbf{Total} &\textbf{} &\textbf{394} &\textbf{38.07} \\
\bottomrule
\end{tabular}
\end{adjustbox}
\label{table-supp:compute}
\end{table}

% \subsection{Error bars}
\textbf{Standard deviation of our experiments (Error Bars).}
We report the standard deviation of MI Attack accuracies for 2 experiment setups: 
$\bullet$ We use $\D_{priv}$ = CelebA \cite{liu2015deep}, $\D_{pub}$ = CelebA \cite{liu2015deep}, $T$ = FaceNet64. 
$\bullet$ We use $\D_{priv}$ = CelebA \cite{liu2015deep}, $\D_{pub}$ = FFHQ \cite{karras2019style}, $T$ = FaceNet64. 
We repeated the entire training and experiments three times. For each trial, we trained T-ACGAN and surrogate models from scratch using different random seeds.
The results are shown in Table \ref{tab:error_bar}.
% \textcolor{red}{\bf [Milad: Good to include the details on how we produce these error bars, e.g., `We run experiments three times and report the average and std of Attack acc. / KNN dist. for these three runs'.]}
%As one can observe, the standard deviations are within an acceptable range.
% We report the results running three times in Table \ref{tab:error_bar}.

\begin{table}[!h]
\centering
\caption{
We report standard deviations for MI Attack accuracies for 2 experiment setups over 3 independent runs. 
The setups include: $\bullet$ We use $\D_{priv}$ = CelebA \cite{liu2015deep}, $\D_{pub}$ = CelebA \cite{liu2015deep}, $T$ = FaceNet64. 
$\bullet$ We use $\D_{priv}$ = CelebA \cite{liu2015deep}, $\D_{pub}$ = FFHQ \cite{karras2019style}, $T$ = FaceNet64. 
We also report the standard deviations for existing SOTA \cite{kahla2022label}.
%We clearly show that our standard deviations are within an acceptable range.
% the entire training and experiments  three times, utilizing the following setup: $T =$ FaceNet64, $\D{priv} =$ CelebA, and $\D_{pub} =$ CelebA/FFHQ. For each trial, we trained T-ACGAN and surrogate models from scratch using a different random seed.
}

% \begin{adjustbox}{width=1.0\columnwidth,center}
\begin{tabular}{lllll}
\hline
\textbf{Setup} &\multicolumn{2}{c}{\textbf{Attack}} & \textbf{Attack acc. $\uparrow$} & \textbf{KNN dt. $\downarrow$} \\ \hline

\multirow{1}{2.8cm}{$T$ \hspace{0.48cm} = FaceNet64  $\D_{priv}$ = CelebA  $\D_{pub}$ \hspace{0.01cm} = CelebA } & \multicolumn{2}{c}{BREPMI} & 74.87	 $\pm $  4.17	& 1286.04	 $\pm $  1.42 \\ \cmidrule{2-5}
& \multirow{3}{*}{LOKT} & $C \circ D$ & 80.80 $\pm $ 4.35  & 1305.97 $\pm $ 		6.50	\\
& &  $S$  & 91.96 $\pm $ 2.62  &	1211.15 $\pm $ 		17.06\\
&  & $S_{en}$  & 93.11 $\pm $ 2.69	  &	1193.16	 $\pm $ 	25.99\\ \hline

\multirow{1}{2.8cm}{$T$ \hspace{0.48cm} = FaceNet64  $\D_{priv}$ = CelebA  $\D_{pub}$ \hspace{0.01cm} = FFHQ} & \multicolumn{2}{c}{BREPMI} &  41.91 $\pm $  	5.09	& 1484.20	$\pm $  13.21 \\ \cmidrule{2-5}
 & \multirow{3}{*}{LOKT} & $C \circ D$ & 44.33	$\pm $  4.25 &  1510.34 $\pm $ 5.07 \\
&  & $S$  & 58.42	$\pm $ 3.61 & 	1439.02 $\pm $ 13.79 \\
& & $S_{en}$  & 62.11	$\pm $  3.66 & 		1426.89 $\pm $ 12.73 \\ \hline
 
\end{tabular}
% \end{adjustbox}
\label{tab:error_bar}
\vspace{-0.5cm}
\end{table}

\end{document}